\documentclass{article}
\usepackage{PRIMEarxiv} 
\usepackage[utf8]{inputenc}
\usepackage[T1]{fontenc}% use 8-bit T1 fonts
\usepackage{hyperref}       % hyperlinks
\usepackage{url}            % simple URL typesetting
\usepackage{booktabs} % professional-quality tables
\usepackage{diagbox}
\usepackage{amsfonts}
\usepackage{nicefrac}
\usepackage{float}
\usepackage{microtype}      % microtypography
\usepackage{lipsum}
\usepackage{amsmath}
 \usepackage{multirow}
\usepackage{bbm}
\usepackage{subcaption}
\usepackage[labelformat=parens,labelsep=quad,skip=3pt]{caption}
\usepackage{graphicx}
\usepackage[superscript,biblabel]{cite}
\usepackage{fancyhdr}       % header
\usepackage{graphicx} % graphics
\graphicspath{{media/}}     % organize your images and other figures under media/ folder
\title{Classification of integers based on residue classes via modern deep learning algorithms
}

\author{
  Da Wu, Jingye Yang \\
  Department of Mathematics \\
  University of Pennsylvania\\
  Philadelphia, PA 19104, USA\\
  \texttt{\{dawu,jingyey\}@math.upenn.edu} \\
  \And
  Mian Umair Ahsan, Kai Wang\thanks{Corresponding author and lead contact} \\
  Department of Pathology \\
  Children's Hospital of Philadelpha \\
  University of Pennsylvania \\
  Philadelphia, PA 19104, USA\\
  \texttt{\{ahsanm1,wangk\}@chop.edu} \\
}

\begin{document}
\maketitle

\begin{abstract}
Judging whether an  integer can be divided by prime numbers such as 2 or 3 may appear trivial to human beings, but can be less straightforward for computers. Here, we tested multiple deep learning architectures and feature engineering approaches on classifying integers  based on their residues when divided by small prime numbers. We found that the ability of classification critically depends on the feature space. We also evaluated Automated Machine Learning (AutoML) platforms from Amazon, Google and Microsoft, and found that they failed on this task without appropriately engineered features. Furthermore, we introduced a method that utilizes linear regression on Fourier series basis vectors, and demonstrated its effectiveness. Finally, we evaluated Large Language Models (LLMs) such as GPT-4, GPT-J, LLaMA and Falcon, and demonstrated their  failures. In conclusion, feature engineering remains an important task to improve performance and increase interpretability of machine-learning models, even in the era of AutoML and LLMs.
\end{abstract}

% keywords can be removed
%\tableofcontents
\section{Introduction}
The task of determining residue class when dividing a given integer, such as $74$ or $243589$, by a prime number, such as $2$ or $3$, remains an interesting and practical problem. In its simplest form, distinguishing whether an integer is odd or even is straightforward for humans. Merely examining the unit digit is sufficient: if it belongs to the set $\lbrace 0, 2, 4, 6, 8\rbrace$, the number is even; otherwise, it is odd.
\par
On the other hand, classifying integers based on their residues when divided by $3$ poses a slightly more difficult challenge. It is well-known that an integer is divisible by $3$ if and only if sum of its numerical digits is divisible by $3$. For example, the number $123$ is divisible by $3$ because $1 + 2 + 3 = 6$, which is divisible by $3$. Conversely, $59$ is not divisible by $3$ as sum of its digits, $5 + 9 = 14$, is not divisible by $3$. Drawing upon this rule, a simple algorithm can be swiftly devised for humans to classify integers based on their residues modulo $3$. When confronted with an arbitrary integer $n$, one can first check if it is divisible by $3$. If it is, the classification is complete. If not, one can try $n+1$ or $n-1$. If either of these numbers is divisible by $3$, the process stops. Otherwise, one can try $n+2$ or $n-2$. At this stage, termination is necessary as there are only three possible residue classes modulo $3$.
\par
Nevertheless, for both cases discussed above, our reliance on mathematical knowledge to design the learning algorithms is absolute. Although no algorithm can achieve satisfactory performance on all possible scenarios based on ``no free lunch theorem'' \cite{wolpert1997no}, Automated Machine Learning (AutoML) has attracted significant attention and demonstrated success in many domain specific problems \cite{karmaker2021automl,he2021automl,truong2019towards}. Multiple academic and commercial implementations of AutoML are now available to help users select the best-performing model for a specific problem. Furthermore, a number of Large Language Models (LLMs) have been developed and popularized \cite{zhao2023survey}, with one well known example being ChatGPT - a powerful chatbot based on LLMs developed by OpenAI \cite{brown2020language,ouyang2022training} . These LLMs showed special abilities that are not present in small-scale language models: for example, in addition to memorizing knowledge, LLMs exhibit reasoning abilities when they are sufficiently large \cite{huang2022towards,saparov2022language,li2022advance,creswell2022selection}. Despite these recent development in the machine-learning space, the question of whether there exists a systematic approach for Turing machines \cite{turing2009computing,button1995computers} to autonomously discern patterns from training data and effectively address the integer classification problem appears to be an intriguing and often overlooked issue. This serves as the central problem of our investigation in the current study. 
\section{Results}
\subsection{Problem setups}
As mentioned above, we are interested in the problem of classifying integers based on their residues mod $p$. Due to practical considerations such as representations and memory limitations, we restrict our sample space to \emph{non-negative integers up to $2^{32}-1$}. For instance, when $p=2$, our parameter space $\mathcal{X}_2=\mathbb{Z}\cap[0,2^{32}-1]$ and our label space $\mathcal{Y}_2=\lbrace 0,1\rbrace$.  Given a set of training data $S_2=\lbrace (x_1,y_1),\ldots, (x_n,y_n), y_i \equiv x_i\mod 2\rbrace\in (\mathcal{X}_2\times\mathcal{Y}_2)^n$,  we want to train a classifier $h_{S_2}:\mathcal{X}_2\mapsto\mathcal{Y}_2$ to predict whether an ``unseen'' integer is odd or even. In this case, we have a binary classification problem. For general $p$, the parameter space $\mathcal{X}_p=\mathbb{Z}\cap[0,2^{32}-1]$ and the label space $\mathcal{Y}_p=\lbrace 0,1,\ldots,p-1\rbrace$. The training set $S_p=\lbrace (x_1,y_1)\ldots,(x_n,y_n):y_i\equiv x_i\mod p \rbrace$ is sampled from $(\mathcal{X}_p\times\mathcal{Y}_p)^n$ and we hope to build a effective classifier $h_{S_p}:\mathcal{X}_p\mapsto\mathcal{Y}_p$ to classify integers based on their residues mod $p$. This time we have a multi-class classification problem.
\par
We primarily focus on the cases when $p=2$ and $3$ but also extend some of the analysis to some other small prime numbers, e.g., $p=7$. 
\subsection{Preparation of datasets}\label{subsec:prep of data}
We uniformly sample integers within the range of $[0,2^{32}-1]$. The specific size of the datasets may vary in different cases and will be specified later. In addition, we also consider the following feature engineering approaches on non-negative integers:
\begin{itemize}
    \item \textbf{Binary representation}: For instance, $4$ is equal to $[0,\ldots,0,1,0,0]$, $2$ is equal to $[0,\ldots,0,0,1,0]$ and $5$ is equal to $[0,\ldots,0,1,0,1]$.
    \item \textbf{Base-three representation}: For example, $3$ is equal to $[0,\ldots,0,1,0]$ and $6$ is equal to $[0,\ldots, 0,2,0]$.
    \item \textbf{One-gram encoding}: We separate the integer into a vector of numerical digits. For instance, $123$ will become $[0,\ldots,0,1,2,3]$
    \item \textbf{Two-gram encoding}: We group two consecutive numerical digits together (with overlap) to form our feature vector. For instance, $1234$ will become $[[0,0],\ldots,[0,0],[0,1],[1,2],[2,3],[3,4]]$.
    \item \textbf{Three-gram encoding}: We group three consecutive numerical digits together (with overlap). For instance, under Three-gram encoding, $1234$ will become $[[0,0,0],\ldots,[0,0,0],[0,0,1],[0,1,2],[1,2,3],[2,3,4]]$.
    \item \textbf{One-gram \& Two-gram encoding combined}: It is the union of One-gram and Two-gram encoding.
    \item \textbf{One-gram \& Two-gram \& Three gram encoding combined}: It is the union of One-gram, Two-gram, and Three-gram encoding.
\end{itemize}
All the above feature engineering processes will be tested on both mod $2$ and mod $3$ cases. In addition to those mentioned above, given the problem nature of mod $3$, we also try the following two feature engineering processes:
\begin{itemize}
    \item \textbf{One-gram encoding$+$its sum}: In addition to One-gram encoding, we also add sum of all of its numerical digits. For instance, $1234$ will become $[0,\ldots,0,1,2,3,4,10]$.
    \item \textbf{One-gram encoding$+$(its sum $\% 3$)}: In this case, $1234$ will become $[0,\ldots,0,1,2,3,4,10 \% 3]=[0,\ldots,0,1,2,3,4,1]$.
\end{itemize}
Since we ran experiments in Python, we adopt conventions of Python: The $[\ldots]$ means list and the $[[\ldots],\ldots,[\ldots]]$ means nested list (a.k.a. matrix). We also pad additional $0$'s or zero lists $[0,\ldots,0]$ on the left to make sure that each feature vector has the same dimension. By doing this, we can covert list-like objects into tensor forms so that we can put them into tensorflow neural networks. 
\subsection{Results from deep neutral networks (DNN)}
\begin{figure}
\begin{subfigure}{5cm}
    \centering\includegraphics[width=4cm]{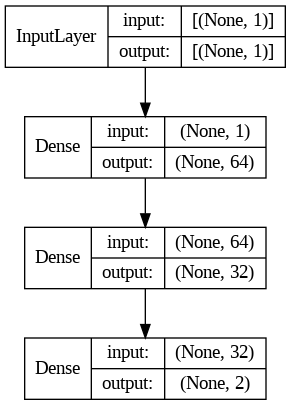}
    \caption{The ANN that we used for training raw data.}
\end{subfigure}
\begin{subfigure}{5cm}
    \centering\includegraphics[width=4cm]{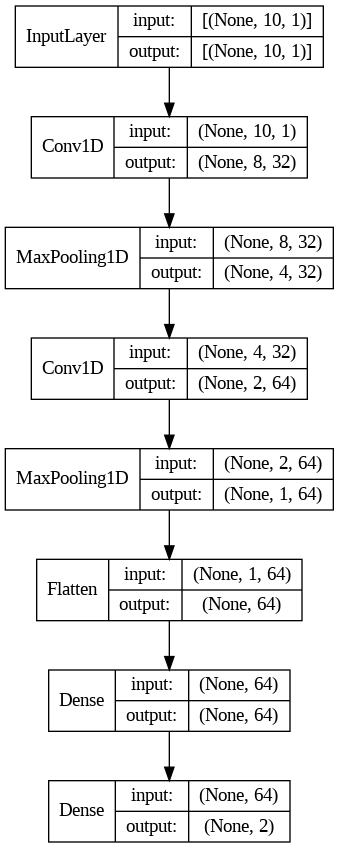}
    \caption{The CNN that we used for training data after One-gram encoding.}
\end{subfigure}
\begin{subfigure}{5cm}
    \centering\includegraphics[width=4cm]{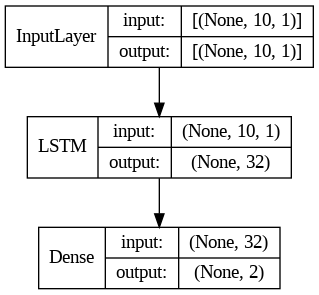}
    \caption{The RNN that we used for training data after One-gram encoding.}
\end{subfigure}
\caption{The deep neural networks considered for mod $2$ and mod $3$ problems}
\label{fig:DNN}
\end{figure}

We first tested on \textit{Artificial Neural Networks} (ANN). The ANN considered here has $64$ neurons in the first layer and $32$ neurons in the second layer. The dimensions of the input and output layer depend on the feature engineering process and the number of training labels,  respectively. The activation functions are the classical ReLU functions, except that for the output layer we use the Sigmoid function. Besides the classical ANNs, we also tested on \textit{Convolutional Neural Network} (CNN) \cite{li2021survey,alzubaidi2021review} and \textit{Recurrent Neural Network} (RNN) \cite{sherstinsky2020fundamentals,yu2019review}. The CNN considered here has two $1$D convolution layers, two max pooling layers, and two dense layers with ReLU activation functions. The RNN considered here has a single LSTM layer and a dense layer with sigmoid activation functions. Finally, we tested on \textit{Bidirectional Encoder Representations from Transformers} (BERT)\cite{DBLP:conf/naacl/DevlinCLT19}, which is a family of masked language models introduced by Google in $2018$. The ANN, CNN and RNN architectures considered in this study are illustrated in Figure \ref{fig:DNN}.
\par 
We first discuss the mod $2$ case. As listed in Table \ref{tab:DNNmod2}, ANN on raw data only gives an accuracy of $\sim 0.5$. It is not surprising that after converting to binary representations, accuracy of $1.000$ can be achieved since we have done mod $2$ operations already during binary transformations. The One, Two and Three gram encodings (or their combined versions) do give reasonable hints to the algorithms and they significantly improve the accuracy to $\sim 0.8$. The BERT model in the end can also achieve an accuracy of $1.000$ due to extensive pre-training processes and the state-of-the-art Transformer architecture.
\par 
In the mod $3$ case, it is expected that base-three representations can achieve an accuracy of $1.000$ even under simple network architectures for the same reason as binary representations. The ``One-gram + (its sum $\% 3$)'' encoding can achieve accuracy of $>0.9$ in all networks tested since it essentially told algorithms everything about the mod $3$ divisibility rule.
\par
It is also  interesting to see that BERT can achieve an accuracy of $1.000$ from ``One-gram+its sum'' encoding whereas it can only have an accuracy of $\sim 0.33$  from ``One-gram'' encoding. In other words, the hint of summing up digits works well with BERT but poorly with other types of deep neural networks. BERT is the only algorithm here that can achieve an accuracy of $1.000$ with ``One-gram + its sum'' encoding in mod $3$ case. The key difference between BERT and other neural networks is that in BERT, all the possible sums of the digits ($99$ potential choices of sums assuming there are $10$ digits maximum when we restricted input space to $2^{32}-1$) are embedded into algorithms with length $768$ each, and therefore we have $768\times 99$ more features in BERT instead of having only one additional feature (sum of digits) in other neural networks. Due to large amounts of pre-training data and better embedding techniques, these $768\times 99$ additional features have already been ``seen and learned'' by BERT and this helps detect the recursive nature of mod $3$ problem. On the contrary, in other neural networks, the sum of digits could be large and therefore completely unseen during training, which leads to inaccuracy.
\par
We found that within Table \ref{tab:DNNmod2} and Table \ref{tab:DNNmod3}, certain outcomes are distinctly surpassing what would be expected from mere random guessing. For instance, in Table \ref{tab:DNNmod2}, ANN (One-gram) attains an accuracy of $0.864$, and ANN (Two-gram) achieves an accuracy of $0.778$. In all these instances, we believe that further escalating complexities of the model in a visible scale, as evidenced by our experimentation, could potentially drive these outcomes to a perfect score of $1.000$. In other words, these cases enjoy the so-called \emph{scaling property} \cite{cohen2021scaling,xu2021convergence}. This fact can also be confirmed during our experimentation on AutoML platforms later. However, in our experiments later, different classes of models were selected and employed, all of which were tree-based algorithms with specific forms of regularization. 
\par
It is important to note, however, that this phenomenon is not replicated across all the other scenarios. For those cases with outcomes equivalent to random guessing (that is, an accuracy of $\sim 0.5$ for modulo 2 and $\sim 0.33$ for modulo 3), even though the well-regarded \textit{Universal Approximation Theorem} \cite{hornik1989multilayer} theoretically assures us of the possibility to perfectly fit training data, it is impractical for individuals to   identify the optimal candidates.
\subsection{Results from AutoML}\label{subsubsec:Auto_ML_results}
We also tested on commonly-used AutoML platforms developed by Google\cite{google_autoML}, Microsoft\cite{microsoft_autoML} and Amazon\cite{amazon_autoML}. To reduce the computing time and save computing power, we uniformly sample $30,000$ non-negative integers from $[0,2^{32}-1]$.
\par
As can be seen in Table \ref{tab:Auto_mod2} and Table \ref{tab:Auto_mod3}, the pre-installed feature engineering algorithms in AutoML pipelines are not effective at all with raw data. This raises an alert that blindly throwing data into AutoML platforms without any feature engineering has certain  (sometimes very high) level of risks; although these AutoML products are extremely powerful, carefully designed and updated constantly by ML experts, they cannot guarantee to deliver an effective model autonomously. It is crucial to apply  domain knowledge to transform data before training.
\par
In the case of mod $3$, as presented in Table \ref{tab:Auto_mod3}, breaking down and/or combining digits without summing them up are not effective. However, summing up all the digits in addition to One-gram encoding can help AutoML platforms deliver classifiers with an accuracy of $1.000$. Note that results of AutoML are reported for the best-possible model from multi-angle considerations (e.g., complexity, interpretations, etc.) and it appears that these AutoML platforms heavily prefer tree-based algorithms, even though deep learning algorithms were also considered. 
\subsection{Results from Fourier series regression}\label{res:fourier}
Next, we proposed a method using Fourier series regressions and tested on mod $3$ and mod $7$ problems. In both cases, an accuracy of $1.000$ can be achieved. This approach proves effective for handling all values of $p$ with a minimal number of training samples, provided that the training dataset size significantly exceeds the value of $p$. To demonstrate this point, this section contains outcomes of both modulo $3$ and modulo $7$ problems, based on a limited dataset of only $200$ samples, divided into $150$ for training and validation ($135$ for training and $15$ for validation) and $50$ for testing. An accuracy of $1.000$ can already be attained for any given test set. Augmenting the sample size by a factor of 10 or 100 does not exhibit any noticeable impact on accuracy and only imparts minimal influence on regression estimations.
\par
Here, our total dataset consists of $200$ uniform samples from $[0,2^{32}-1]$, among which $150$ samples are used to estimate regression coefficients and the remaining $50$ ones are used for testing purposes. Fix integer $p=3$ or $7$. Suppose $X = [x_1,\ldots,x_n ]^T$ is our training set and let $Y_p=[y_1,\ldots,y_n]^T$ be the vector of training labels, i.e., $y_i \equiv x_i\mod p$, where $y_i\in \lbrace 0,1\ldots,p-1\rbrace$ and $p\in \lbrace 3,7\rbrace$. For each $j = 1,2,\ldots,[p/2]$, let 
\begin{equation}\label{fourier_sine}
    \sin\left(\frac{2\pi j}{p}X \right)=\left[\sin\left(\frac{2\pi j}{p}\cdot x_1 \right),\ldots, \sin\left(\frac{2\pi j}{p}\cdot x_n \right) \right]^T
\end{equation}
and 
\begin{equation}\label{fourier_cosine}
    \cos\left(\frac{2\pi j}{p}X \right)=\left[\cos\left(\frac{2\pi j}{p}\cdot x_1 \right),\ldots, \cos\left(\frac{2\pi j}{p}\cdot x_n \right) \right]^T
\end{equation}
be vectors of Fourier series basis and we consider the following linear regression
\begin{equation}\label{eqn:fourier_series_reg_3_7}
    Y_p = \gamma +\sum_{j=1}^{[p/2]}\left(\alpha_j \sin\left(\frac{2\pi j}{p}X\right) + \beta_j \cos\left(\frac{2\pi j}{p}X\right)\right)+\varepsilon,
\end{equation}
where $\varepsilon$ denote the vector of standard Gaussian noise. All the coefficients in (\ref{eqn:fourier_series_reg_3_7}) are estimated by classical \textit{ordinary least square} (OLS) regression.
\par
Generally speaking, linear regression model is not a good candidate for classification problems due to a number of reasons, one of them being that the output values are continuous instead of categorical. However, in our case, due to the large number of training samples, predicted values on the testing set are very close to integers (see Table \ref{tab:F3_value} and Table \ref{tab:F7_value}) so that we can round them to the nearest integer if their distance is within $10^{-5}$.
\subsubsection{Regression estimates of mod $3$}
\begin{figure}[ht]
    \centering
    \includegraphics[width=1\textwidth]{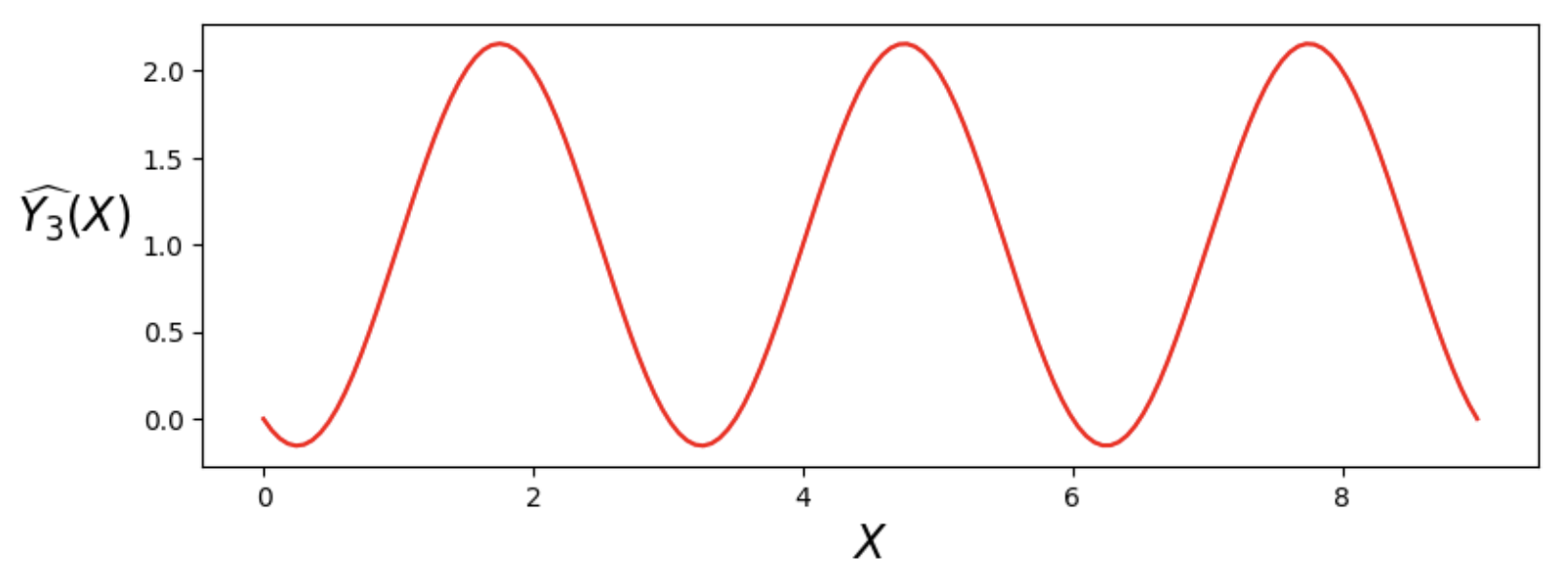}
    \caption{Plot of the regression estimate $\widehat{Y_3}(X)$ (\ref{eqn:reg results on mod3}).}
    \label{fig:F3}
\end{figure}

By using \texttt{LinearRegression} package in \texttt{sklearn} of Python, we have the following regression coefficients estimate: 
\begin{align}\label{eqn:reg results on mod3}
     \widehat{Y_3}(X) = 0.9999999556293363 + (-0.57735)\cdot\sin\left(\frac{2\pi}{3}X \right)+ (-1.00000)\cdot\cos\left(\frac{2\pi}{3}X \right)
\end{align}
with $R^2$ value being $0.9999999999999453$ and all coefficients being statistically significant. The $R^2$ value is computed under train-validation split ratio $0.1$. The plot of regression estimate (\ref{eqn:reg results on mod3}) and its values at integer points, covering three periods, are recorded in Figure \ref{fig:F3} and Table \ref{tab:F3_value} respectively. In the mod $3$ case, accuracy $1.000$ can be achieved. We also want to emphasize that the joint presence of both sines and cosines is needed and if we, for instance, remove all the cosines and only keep the sines in (\ref{eqn:fourier_series_reg_3_7}), then $R^2$-value is only  $0.2533$ and accuracy is $0.3206$ with the same dataset. On the contrary, we can also add more pairs of sine and cosine, e.g., $j = 1,\ldots p-1$, in our Fourier series basis (\ref{fourier_sine}) and (\ref{fourier_cosine}). This will also give us a satisfying linear regression estimate (accuracy $1.000$) with ``less overshoot and undershoot'' compared to (\ref{eqn:fourier_series_reg_3_7}). However, the regression table will give the potential warning of \textit{multicollinearity} \cite{alin2010multicollinearity,mansfield1982detecting,daoud2017multicollinearity}, indicating that too many features were added, which may cause instability of regression coefficients estimates in some large $p$ cases. Therefore, our linear regression model (\ref{eqn:fourier_series_reg_3_7}) is \emph{optimal}.
\subsubsection{Regression estimates of mod $7$}
\begin{figure}
    \centering
    \includegraphics[width=1\textwidth]{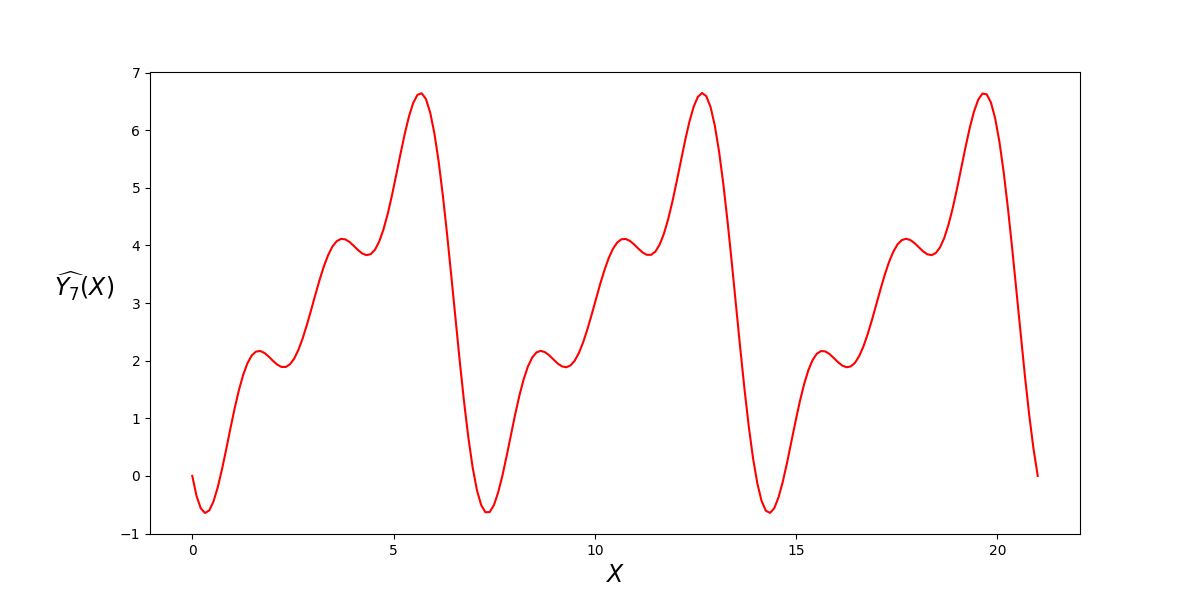}
    \caption{Plot of the regression estimate $\widehat{Y_7}(X)$ (\ref{eqn:reg results on mod7}).}
    \label{fig:F7}
\end{figure}

In the case of mod $7$, we have the following coefficients estimates: 
\begin{align}\label{eqn:reg results on mod7}
    \begin{split}
        \widehat{Y_7}(X) = 3.0000000310122816 & + (-2.076521)\cdot \sin\left(\frac{2\pi}{7}X \right) +(-1.000000)\cdot \cos\left(\frac{2\pi}{7}X \right)\\
        & + (-0.797473)\cdot \sin\left(\frac{4\pi}{7}X\right)+ (-1.000000)\cdot\cos\left(\frac{4\pi}{7}X \right)\\
        & + (-0.228243)\cdot \sin\left(\frac{6\pi}{7}X\right)+ (-1.000000)\cdot\cos\left(\frac{6\pi}{7}X \right),
    \end{split}
\end{align}
with $R^2$-value being $0.9999999999999203$ and all coefficients being statistically significant. Again, accuracy of $1.000$ can be achieved on the testing set. The plot of regression estimate (\ref{eqn:reg results on mod7}) and its values at integer points, covering two periods, are recorded in Figure \ref{fig:F7} and Table \ref{tab:F7_value}, respectively.
\subsection{Results from Large Language Models (LLMs)}
Next, we conducted tests to assess the proficiency of open-source LLMs, specifically GPT-J-6B \cite{mesh-transformer-jax,gpt-j}, LLaMA-7B \cite{touvron2023llama}, and Falcon-40B \cite{falcon_40b_huggingface}, in their understanding of divisibility rules.  Specifically, for each prime $p$ up to $31$, we tried the following two prompts:
\begin{itemize}
    \item \textbf{P1}: ``There are various mathematical rules to check if an integer is divisible by $p$, for instance''
    \item \textbf{P2}: ``How to check if an integer is divisible by $p$, for instance we can''
\end{itemize}
We utilized these two prompts to better steer the open-source models towards producing mathematical answers as opposed to algorithmic ones. Additionally, we included ``for instance$\ldots$'' since open-source models are more suitable for completing sentences rather than providing direct answers to questions. 
\par
The comprehensive responses are presented in \textbf{Note S1-S3}. Within this context, we manually assess the accuracy and informativeness of responses generated by those models. The elaborated outcomes are recorded in Table \ref{tab:res_LLM}. For ``Correct'', we refer to the \emph{mathematical accuracy of the provided answers}. In terms of ``Informative'', we assess \emph{whether the models effectively articulate the divisibility rule in a lucid and cohesive manner}. As an example, when the model attempts to employ the modulo operator ($\%$) - while its mathematical accuracy is unquestionable - it ultimately lacks any informative value. The data presented in Table \ref{tab:res_LLM} demonstrates the inadequacy of these open-source LLMs for primes beginning at $7$. Even when dealing with smaller primes, they consistently produce incorrect and uninformative replies.
\par 
Furthermore, we assessed ChatGPT's \url{https://openai.com/blog/chatgpt} closed-source implementation utilizing reinforcement learning on the knowledge of divisibility rules. All outcomes can be entirely replicated, and to ensure alignment with open-source models, we also incorporate ChatGPT's evaluations in Table \ref{tab:res_LLM}. Specifically, we used the following prompt:
\begin{itemize}
    \item \textbf{P}: ``How to check if an integer is divisible by $p$?''.
\end{itemize}
Regarding ChatGPT, utilizing the aforementioned prompts (\textbf{P1} and \textbf{P2}) is unnecessary, as it possesses a strong capability to comprehend the genuine intention behind the prompt $\textbf{P}$. The comprehensive results are furnished in \textbf{Table S1-S6}. Notable enhancements are observed compared to open-source LLMs. Nevertheless, beginning at $23$, ChatGPT resorts to using the modulo operator $\%$ to tackle the issue, resulting in completely uninformative answers. As a result, its adequacy for larger primes still lags behind. 
\par
Finally, we tested the latest GPT-4 with code interpreter (distinct from the aforementioned ChatGPT) on its capabilities of designing deep neural networks to address the integer divisibility problems. Note that as of now, access to GPT-4 is exclusively offered to subscribers of ChatGPT plus. We showcased its code in the \textbf{Note S4}. While correct, it does not possess the capability to formulate an efficient algorithm, which encompasses identifying appropriate feature engineering techniques and selecting the optimal ML/DL algorithms.
\par 
Hence, it can be deduced that  current state-of-the-art LLMs do not possess the required capability to offer dependable and accurate information regarding divisibility rules.
\section{Discussion}
In our current study, we conducted extensive experiments to delve into classifications of large finite integers, specifically those up to $2^{32}$, based on their residues when divided by small prime numbers. Our investigation involved testing various deep neural network architectures and employing diverse feature engineering approaches. The obtained results were both intuitive and straightforward to interpret.
\par
An important observation emerged throughout our analysis is that, regardless of complexities of network structures or specific neural network frameworks used, the performance of our classification task was highly reliant on the feature space provided to deep learning models. This discovery remained consistent not only for neural network architectures but also when evaluating AutoML pipelines.
\par
Feature engineering is often a challenging and non-intuitive process in practical scenarios, requiring extensive trial-and-error iterations \cite{heaton2016empirical,ozdemir2018feature,hermann2020shapes}. In addition to directly engineering features on training samples, there are other avenues where domain expertise can be leveraged to enhance the quality of classifiers. Inspired by the \textit{recurring pattern} exhibited by the modulus function, we devised a simple method that utilizes linear regression with Fourier series basis vectors to capture and understand its periodic behavior. This method exhibited exceptional performance, achieving a perfect accuracy of $1.000$ for all modulus $p$ problems, even when $p$ is not necessarily prime. Furthermore, our proposed approach offers advantages such as minimal training size, reduced time complexity, and improved interpretability of the model, outperforming all the other state-of-the-art ML/DL models in these aspects.
\par
To expand our evaluation further, we also examined the performance of GPT-J-6B, LLaMA-7B, and Falcon-40B concerning divisibility by primes up to $31$. Regrettably, our investigation uncovered that these open-source LLMs exhibited a tendency to produce inaccurate and uninformative replies even when primes $p$ are small. Additionally, we conducted tests on closed-source ChatGPT and observed a relatively improved performance compared to the aforementioned open-source LLMs. However, it is important to note that ChatGPT still demonstrated some instances of erroneous information, particularly when dealing with larger prime number $p$. For $p\geq 23$, it cannot provide any informative answers at all. Finally, we tested the latest GPT-4 with code interpreter on its capabilities of designing an effective neural network to address the mod $p$ problem and found that it still lacks the competence to do so.  
\par
In conclusion, our study emphasizes the ongoing importance of feature engineering in the AutoML and LLM era. We demonstrated that performances of deep learning models are heavily reliant on carefully engineered features. Without appropriate feature engineering, it is impossible to enhance performance merely by increasing complexities of algorithms on a feasible level or enlarging sizes of training datasets; despite the renowned Universal Approximation Theorem \cite{hornik1989multilayer} offering a theoretical assurance that achieving a perfect fit to the training data is possible, it is impractical for individuals to effectively discover the desired candidates in practice. Moreover, our proposed method employing linear regression on Fourier series basis vectors showcased exceptional accuracy, lower time complexity, and enhanced interpretability. Finally, we caution against relying on LLMs for divisibility by large primes, as our findings indicated that they may provide inaccurate information in these cases.
\section{Experimental Procedures}
\subsection{Resource availability}
\subsubsection{Lead contact}
Further information and requests for data should be directed to and will be fulfilled by the lead contact, Dr. Kai Wang (wangk@chop.edu).
\subsubsection{Materials availability}
This study did not generate new unique materials.
\subsubsection{Data and code availability}
All the publicly available code for producing training data, conducting testing using various feature engineering approaches, and implementing distinct deep learning algorithms outlined in Table \ref{tab:DNNmod2} and Table \ref{tab:DNNmod3}, as well as the Fourier series regression discussed in Section \ref{res:fourier}, can be accessed at \cite{da_wu_2023_8327655} (\href{https://doi.org/10.5281/zenodo.8327655}{DOI:10.5281/zenodo.8327655}). 
\subsection{Deep neutral networks}\label{sec:DNN methods}
The ANN considered in this study has $64$ neurons in the first layer and $32$ neurons in the second layer. The dimension of the input and output layer depend on what feature engineering process we are using and how many training labels there are. The activation functions are the classical ReLU functions, except that for the output layer we use the Sigmoid function. Notice that this is the only ANN considered in this study and we will not compare the results with other more complex ANNs. This consideration is due to the facts that (1) the above neural network is sufficient for us to demonstrate the importance of feature engineering, and (2) the more complex networks will be tested on the commonly used AutoML platforms.
\par
Besides the classical ANNs, we also considered CNN and RNN. CNN is a well-known deep learning algorithm inspired by the natural vision perception mechanisms of the living creatures. The modern framework of CNN was established by LeCun \emph{et al.} in \cite{lecunetal90} and later improved  \cite{Lecunetal98}. For recent advances in CNN and its applications, see  \cite{ADCNN18} and the references therein. The RNN architecture was mainly designed to overcome the issue of ``limited context memory'' in \cite{Bengio2003ANP}: only a fixed number of previous words can be taken into account to predict the next word. In RNN, the context length was extended to indefinite size, which can handle arbitrary context lengths. See \cite{RNN14} for the review on the RNN model and its applications on the statistical language modeling. The CNN we used has two $1$D convolution layers, two max pooling layers, and two dense layers with ReLU activation functions. The RNN we used has a single LSTM layer and a dense layer with Sigmoid activation functions. We used 10 epochs with a batch size of $32$ during training.
\par
Finally, we tested on \textit{Bidirectional Encoder Representations from Transformers} (BERT), which is a family of masked language models introduced by Google in $2018$ \cite{DBLP:conf/naacl/DevlinCLT19}. It is based on the Transformer encoders, and was pre-trained simultaneously on two tasks: \textit{masked language modeling} and \textit{next sentence prediction}. After pre-training, it can be \textit{fine-tuned} on smaller datasets to optimize its performance on specific tasks, including text classifications, language inference, etc. See also \cite{trans_survey,survey_on_vision_trans,trans_in_vision,case_study_on_bert} for more discussions on the BERT model and the Transformer architecture. Here we tested the BERT model on both mod $2$ and $3$ problems via the \texttt{BertForSequenceClassification} package in Python. We used $10$ epochs with a batch size of $64$ and a learning rate of $10^{-6}$.
\subsection{Automated Machine Learning}
We considered three commonly-used AutoML platforms: Microsoft Azure ML Studio \cite{microsoft_autoML}, Google Cloud Vertex AI \cite{google_autoML}, and Amazon AWS Sagemaker \cite{amazon_autoML}. All of them are easily accessible online with limited free credits. One can upload the datasets in the ``.csv'' format and pre-specify the target value, type of tasks (classification, regression, NLP, etc.), primary metric of evaluation, etc.
\subsection{Fourier series regressions}
In this section, we described our Fourier series regression method in details. Fix an integer $p\geq 2$. Let $X = [x_1,\ldots,x_n ]^T$ be our training set and let $Y_p=[y_1,\ldots,y_n]^T$ be the vector of training labels, i.e., $y_i \equiv x_i\mod p$ where $y_i\in \lbrace 0,1\ldots,p-1\rbrace$. For each $x_i$ in $X$, consider the following Fourier series basis:
\begin{equation}\label{eqn:fourier_series_basis}
   \left\lbrace  \sin\left(\frac{2\pi j}{p}\cdot x_i \right), \cos\left(\frac{2\pi j}{p}\cdot x_i \right),\ \text{where}\  j = 1,2,\ldots, [p/2] \right\rbrace.
\end{equation}
For each $j$, let 
\begin{equation*}
    \sin\left(\frac{2\pi j}{p}X \right)=\left[\sin\left(\frac{2\pi j}{p}\cdot x_1 \right),\ldots, \sin\left(\frac{2\pi j}{p}\cdot x_n \right) \right]^T
\end{equation*}
and 
\begin{equation*}
    \cos\left(\frac{2\pi j}{p}X \right)=\left[\cos\left(\frac{2\pi j}{p}\cdot x_1 \right),\ldots, \cos\left(\frac{2\pi j}{p}\cdot x_n \right) \right]^T
\end{equation*}
be the vectors of Fourier series basis and consider the following linear regression equation:
\begin{equation}\label{eqn:fourier_series_reg}
    Y_p =
        \gamma +\sum_{j=1}^{[p/2]}\left(\alpha_j \sin\left(\frac{2\pi j}{p}X\right) + \beta_j \cos\left(\frac{2\pi j}{p}X\right)\right)+\varepsilon,
\end{equation}
where $\varepsilon$ denotes the vector of standard Gaussian noise. All the coefficients in (\ref{eqn:fourier_series_reg}) are estimated by the classical \textit{ordinary least square} (OLS) method. For prediction, we round the predicted values from linear regression (\ref{eqn:fourier_series_reg}) to the nearest integer if their distance is within $10^{-5}$.
\subsection{Large Language Models (LLMs)}\label{subsec:methods LLM}
Finally, we tested the knowledge of GPT-J-6B \cite{mesh-transformer-jax}, LLaMA-7B\cite{touvron2023llama}, and Falcon-40B \cite{tii_falcon_40B} on divisibility rules of prime numbers. These models are all open-sourced and can be easily implemented using hugging face API\cite{gpt_j_huggingface,llama_huggingface,falcon_40b_huggingface}.
\par
For demonstration purposes, we conducted tests on several scenarios where $p$ took on the primes up to $31$.  For each prime $p$, we employed the following two prompts for each of these three models, ensuring consistency across the evaluations:
\begin{enumerate}
    \item ``How to check if an integer is divisible by $p$, for instance we can''
    \item ``There are various mathematical rules to check if an integer is divisible by $p$, for instance''
\end{enumerate}
The maximum token lengths for GPT-J-6B, LLaMA-7B, and Falcon-40B are set to be $100,160,$ and $200$ respectively. For ChatGPT\cite{chat_gpt}, we used ``How to check if an integer is divisible by $p$?'' as the prompt. For GPT-4, we used ``Can you design a deep learning algorithm to determine if a number can be divided by
$p$?''
\section{Acknowledgments}
This project originated from a discussion between Umair Ahsan (CHOP), Qian Liu (UNLV) and Kai Wang (CHOP/Penn) during the peak of COVID-19. We are grateful to Dr. Xin Sun (Penn), Dr. Herman Gluck (Penn) and Penn Data Driven Discovery Initiative for their support of pursuing this project. We thank Wang lab members for helpful discussions and comments on the methods and results.
\section{Author contributions}
Conceptualization, D.W. and K.W.; Methodology, D.W., J.Y., M.U.A, and K.W.; Investigation, D.W., J.Y., M.U.A, and K.W.; Writing-Original Draft, D.W. and K.W.; Writing–Review \& Editing, D.W., J.Y., and K.W.; Funding Acquisition, K.W.; Supervision, K.W.
\section{Declaration of interests}
The authors declare no competing interests.
%Bibliography
\bibliographystyle{unsrt}  
\bibliography{references}  
\newpage
\begin{table}[ht]
    \centering
    \begin{tabular}{|l||c|c|}
 \hline
 \multicolumn{3}{|c|}{Classification on divisibility by $2$ via Deep Neural Networks} \\
 \hline
Algorithms (Feature engineering) & Mean & Standard Deviation\\
 \hline
 Artificial Neutral Network (ANN)    & $0.501$    & $0.003$ \\
 \hline
 ANN (Binary representation) & $1.000$    & $0.000$ \\
 \hline
 ANN (Base-three representation) & $0.537$ & $0.002$ \\
 \hline
 ANN (One-gram) & $0.864$ & $0.024$\\
 \hline
 ANN (Two-gram) & $0.778$ & $0.048$ \\
 \hline
 ANN (Three-gram) & $0.799$ & $0.002$\\
 \hline
 ANN (One $\&$ Two-gram combined) & $0.786$ & $0.028$\\ 
 \hline
 ANN (One $\&$ Two $\&$ Three-gram combined) & $0.781$ & $0.023$\\
 \hline
 Convolutional Neural Network (One-gram) & $0.921$ & $0.042$ \\
 \hline
 Recurrent Neural Network (One-gram) & $0.740$ & $0.026$ \\
 \hline
 Bidirectional Encoder Representations from Transformers (BERT) (One-gram) & $1.000$ & $0.000$ \\ 
\hline
\end{tabular}
    \caption{Results on mod $2$ via deep neural networks. Means and standard deviations are calculated with three uniformly-sampled, equal-size training sets.}
    \label{tab:DNNmod2}
\end{table}
\begin{table}[ht]
    \centering
    \begin{tabular}{ |l||c|c|  }
 \hline
 \multicolumn{3}{|c|}{Classification on divisibility by $3$ via Deep Neural Networks} \\
 \hline
 Algorithms (Feature engineering)  & Mean & Standard Deviation\\
 \hline
 Artificial Neutral Network (ANN)    & $0.334$    & $0.002$ \\
 \hline
 ANN (Binary representation) & $0.396$    & $0.002$ \\
 \hline
 ANN (Base-three representation) & $1.000$ & $0.000$ \\
 \hline
 ANN (One-gram) & $0.343$ & $0.003$\\
 \hline
 ANN (One-gram+its sum)& $0.335$ & $0.002$ \\
 \hline
 ANN (One-gram+(its sum$\% 3$))& $1.000$ & $0.000$ \\
 \hline
 ANN (Two-gram) & $0.334$ & $0.002$\\
 \hline
 ANN (Three-gram) & $0.333$ & $0.001$\\
 \hline
 ANN (One $\&$ Two-gram combined) & $0.332$ & $0.004$\\ 
 \hline
 ANN (One $\&$ Two $\&$ Three-gram combined) & $0.334$ & $0.007$\\
 \hline
 Convolutional Neural Network (CNN) (One-gram) & $0.340$ & $0.002$ \\
 \hline
 CNN (One-gram+its sum)& $0.330$ & $0.002$ \\
 \hline
 CNN (One-gram+(its sum$\% 3$))& $0.936$ & $0.001$ \\
 \hline
 Recurrent Neural Network (RNN) (One-gram) & $0.330$ & $0.003$ \\
 \hline
 RNN (One-gram+its sum)& $0.333$ & $0.001$ \\
 \hline
 RNN (One-gram+(its sum$\% 3$))& $0.934$ & $0.001$\\
 \hline
 Bidirectional Encoder Representations from Transformers (BERT) (One-gram) & $0.336$ & $0.005$ \\
 \hline
 BERT (One-gram + its sum) & $1.000$ & $0.000$ \\
 \hline
 BERT (One-gram + (its sum $\% 3$)) & $1.000$ & $0.000$ \\
\hline
\end{tabular}
    \caption{Results on mod $3$ via deep neural networks. Means and standard deviations are calculated with three uniformly-sampled, equal-size training sets.}
    \label{tab:DNNmod3}
\end{table}
\begin{table}[ht]
    \centering
    \begin{tabular}{ |l||c c|c c|c c|  }
 \hline
\multirow{2}{*}{Feature engineering} & \multicolumn{2}{c|}{Microsoft Azure ML} & \multicolumn{2}{c|}{Google Cloud Vertex AI} &\multicolumn{2}{c|}{Amazon Sagemaker}\\
 \cline{2-7}
  & Accuracy & Model & Accuracy & Model & Accuracy & Model\\
 \hline
 \hline
 No feature engineering   & $0.504$ & RandomForest  &  $0.495$ & N/A & $0.508$ & WeightedEnsemble\\
 \hline
 Binary representation  & $1.000$  & XGBoost &  $1.000$ & N/A & $1.000$ & LightGBM\\
 \hline
 Base-three representation  &  $0.508$ & RandomForest &  $0.506$ & N/A & $0.514$ & WeightedEnsemble\\
 \hline
 One-gram  & $1.000$  & XGBoost &  $1.000$ & N/A & $1.000$ & LightGBM\\
 \hline
 Two-gram  &  $1.000$ & LightGBM &  $1.000$ & N/A & $1.000$ & LightGBM\\
 \hline
 Three-gram  &  $1.000$ & XGBoost &  $1.000$ & N/A & $1.000$ & LightGBM\\
 \hline
One $\&$ Two-gram & $1.000$ &  XGBoost & $1.000$ & N/A & $1.000$ & LightGBM\\
\hline
One $\&$ Two $\&$ Three-gram  & $1.000$ & RandomForest  & $1.000$  & N/A & $1.000$ & LightGBM\\
 \hline
\end{tabular}
    \caption{Results on mod $2$ via AutoML. For Miscrosoft and Amazon, we report the best possible model and their respective performance on test test. The Google AutoML platform did not report the specific model and only reported testing statistics.}
    \label{tab:Auto_mod2}
\end{table}
\begin{table}[ht]
    \centering
    \begin{tabular}{ |l||c c|c c|c c|  }
 \hline
\multirow{2}{*}{Feature engineering} & \multicolumn{2}{c|}{Microsoft Azure ML} & \multicolumn{2}{c|}{Google Cloud Vertex AI} &\multicolumn{2}{c|}{Amazon Sagemaker}\\
 \cline{2-7}
  & Accuracy & Model & Accuracy & Model & Accuracy & Model \\
 \hline
 \hline
 No feature engineering   & $0.357$  & RandomForest & $0.352$ & N/A & $0.345$ &  WeightedEnsemble\\
 \hline
 Binary representation  &  $0.361$ & RandomForest &   $0.342$ & N/A & $0.345$ &  WeightedEnsemble\\
 \hline
 Base-three representation  & $1.000$  & XGBoost & $1.000$ & N/A  & $1.000$ & LightGBM\\
 \hline
 One-gram &  $0.352$ & RandomForest &  $0.352$ & N/A & $0.348$ &  WeightedEnsemble\\
 \hline
 One-gram+its sum& $1.000$ & XGBoost & $1.000$ & N/A & $1.000$ & LightGBM\\
 \hline
 One-gram+(its sum$\% 3$)& $1.000$ & LightGBM & $1.000$ & N/A & $1.000$ & LightGBM\\ 
 \hline
 Two-gram &  $0.359$ & RandomForest &  $0.353$ & N/A & $0.348$ &  WeightedEnsemble\\
 \hline
 Three-gram &  $0.346$ & RandomForest &  $0.349$ & N/A & $0.351$ &  WeightedEnsemble\\
 \hline
 One $\&$ Two-gram&  $0.348$ & RandomForest & $0.351$ & N/A & $0.349$ &  WeightedEnsemble\\
 \hline
 One $\&$ Two $\&$ Three-gram & $0.346$ & XGBoost & $0.341$ & N/A & $0.345$ &  WeightedEnsemble\\
 \hline
\end{tabular}
    \caption{Results on mod $3$ via AutoML. For Miscrosoft and Amazon, we report the best possible model and their respective performance on test test. The Google AutoML platform did not report the specific model and only reported testing statistics.}
    \label{tab:Auto_mod3}
\end{table}
\begin{table}[ht]
    \centering
    \begin{tabular}{|c|c|c|} 
 \hline
 $X$  & $\widehat{Y_3}(X)$  & $X\mod 3$ \\
\hline
 $0$ & $-5.473399511402022\cdot 10^{-14}$ & $0$ \\
 \hline
 $1$ &  $1.0000002331249993$ & $1$ \\
 \hline
 $2$ & $1.9999997668748912$ & $2$ \\
 \hline
 $3$ &  $-5.46229728115577 \cdot 10^{-14}$ & $0$ \\
 \hline
 $4$ &  $1.0000002331249984$ & $1$ \\
 \hline
 $5$ &  $1.999999766874892$ & $2$ \\
 \hline
 $6$ & $-5.440092820663267\cdot 10^{-14}$ & $0$ \\
 \hline
 $7$ & $1.0000002331249984$ & $1$ \\
 \hline
 $8$ &  $1.999999766874892$ & $2$ \\
 \hline
 $9$ &  $-5.4289905904170155\cdot 10^{-14}$ & $0$ \\
 \hline
 \end{tabular}
    \caption{Values of the regression estimate $\widehat{Y_3}(X)$ (\ref{eqn:reg results on mod3}) at integer points.}
    \label{tab:F3_value}
\end{table}
\begin{table}[]
    \centering
    \begin{tabular}{|c|c|c|} 
 \hline
 $X$  & $\widehat{Y_7}(X)$  & $X\mod 7$ \\
\hline
 $0$ & $3.1012281631603855\cdot 10^{-8}$ & $0$ \\
 \hline
 $1$ &  $1.0000009260275013$ & $1$ \\
 \hline
 $2$ & $1.9999998780188157$ & $2$ \\
 \hline
 $3$ &  $3.000000361534261$ & $3$ \\
 \hline
 $4$ &  $3.9999997004903016$ & $4$ \\
 \hline
 $5$ &  $5.000000184005747$ & $5$ \\
 \hline
 $6$ & $5.999999135997062$ & $6$ \\
 \hline
 $7$ & $3.101228274182688\cdot 10^{-8}$ & $0$ \\
 \hline
 $8$ &  $1.000000926027501$ & $1$ \\
 \hline
 $9$ &  $1.9999998780188162$ & $2$ \\
 \hline
 $10$ &  $3.000000361534261$ & $3$ \\
 \hline
 $11$ &  $3.999999700490302$ & $4$ \\
 \hline
 $12$ &  $5.000000184005747$ & $5$ \\
 \hline
 $13$ &  $5.999999135997064$ & $6$ \\
 \hline
 $14$ &  $3.1012283852049904\cdot 10^{-8}$ & $0$ \\
 \hline
 $15$ &  $1.0000009260274998$ & $1$ \\
 \hline
 $16$ &  $1.9999998780188164$ & $2$ \\
 \hline
 \end{tabular}
    \caption{Values of the regression estimate $\widehat{Y_7}(X)$ (\ref{eqn:reg results on mod7}) at integer points}
    \label{tab:F7_value}
\end{table}
\begin{table}[ht]
    \centering
    \begin{tabular}{|c|c c | c c|c c | c c|c c | c c|c |c|}
    \hline
      Prime $p$  & \multicolumn{4}{c|}{GPT-J-6B} & \multicolumn{4}{c|}{LLaMA-7B} &\multicolumn{4}{c|}{Falcon-40B} &\multicolumn{2}{c|}{ChatGPT-175B}\\
      \cline{2-15}
      & \multicolumn{2}{c|}{Correct} & \multicolumn{2}{c|}{Informative} & \multicolumn{2}{c|}{Correct} & \multicolumn{2}{c|}{Informative} & \multicolumn{2}{c|}{Correct} & \multicolumn{2}{c|}{Informative} & Correct & Informative\\
      \cline{2-15}
      & \textbf{P1} & \textbf{P2} & \textbf{P1} & \textbf{P2} & \textbf{P1} & \textbf{P2} & \textbf{P1} & \textbf{P2} & \textbf{P1} & \textbf{P2} & \textbf{P1} & \textbf{P2} & \textbf{P} & \textbf{P}\\
      \cline{2-15}
      \hline
        2 & No & No & Yes & No & No & Yes & No & No & No & No & No & No &Yes & Yes\\
        \hline
        3 & No & No & No & No & Yes & Yes & Yes & Yes & Yes & Yes & Yes & No&Yes & Yes\\
        \hline
        5 & No & No & No & No & Yes & Yes & Yes & Yes & Yes & Yes & No & No&Yes & Yes\\
        \hline
        7 & Yes & No & Yes & No & No & No & No & No & No & No & No & No&Yes & Yes\\
        \hline
        11 & No & No & No & No & No & No & No & No & No & No & No & No&Yes & Yes\\
        \hline
        13 & No & No & No & No & No & No & No & No & No & No & No & No&Yes & Yes\\
        \hline
        17 & No & No & No & No & No & No & No & No & No & No & No & No&Yes & Yes\\
        \hline
        19 & No & No & No & No & No & No & No & No & No & No & No & No& No & Yes\\
        \hline
        23 & No & No & No & No & No & No & No & No & No & No & No & No& Yes & No\\
        \hline
        29 & No & No & No & No & No & No & No & No & No & No & No & No& Yes & No\\
        \hline
        31 & No & No & No & No & No & No & No & No & No & No & No & No& Yes & No\\
        \hline
    \end{tabular}
    \caption{The proficiency of LLMs in providing both accurate mathematical solutions and informative explanations regarding divisibility rules by prime numbers up to $31$. The detailed responses are documented in \textbf{Note S1-S3} and \textbf{Table S1-S6}.}
    \label{tab:res_LLM}
\end{table}
\end{document}

\maketitle
\section{Response from Large Language Models regarding divisibility rules}
In this section, we document extensive answers provided by GPT-J-6B, LLaMA-7B, Falcon-40B, and ChatGPT regarding divisibility rules by primes up to $31$. Additionally, we evaluate the proficiencies of GPT-4 (with code interpreter) in designing algorithms for the mod $17$ problem.
\subsection{Results from GPT-J-6B}
When $p=2$, we have the following two responses:
\begin{itemize}
    \item \textbf{Prompt}: \textit{There are various mathematical rules to check if an integer is divisible by 2, for instance} \\
    \textbf{GPT-J}: There are various mathematical rules to check if an integer is divisible by 2, for instance.  For the numbers in your list, only the 1, 3, 9, and 17 are divisible by 2.  For the others, divide by 2 to check that they work.\textbackslash n\textbackslash nA:\textbackslash n \textbackslash n This question was posted a while ago. This answer works for numbers in any base! Here's the full code that does what I posted before:\textbackslash ndef divisibility(n
    \item \textbf{Prompt}: \textit{How to check if an integer is divisible by 2, for instance we can} \\
    \textbf{GPT-J}: How to check if an integer is divisible by 2, for instance we can create some code that will check if a number is even or not and if it is even it will return true and if it is odd then it returns false and we can make code to check if a number in divisible by 3\textbackslash n\textbackslash nhow to check if an number is divisible by 2, for instance we can create some code that will check if a number is even or not and if it is even it will
\end{itemize}
When $p=3$, we have the following two responses:
\begin{itemize}
    \item \textbf{Prompt}: \textit{There are various mathematical rules to check if an integer is divisible by $3$, for instance} \\
    \textbf{GPT-J}: There are various mathematical rules to check if an integer is divisible by 3, for instance.\textbackslash nAnother one to check if an integer is divisible by 5 or by 7.\textbackslash n \textbackslash nIn any case, these rules will be implemented differently in different programming languages.\textbackslash nTo check if an integer is divisible by 3, you need to look and see if it's divisible by 3, if it is, it has to be divisible by 3. \textbackslash nNot at all the same in
\item \textbf{Prompt}: \textit{How to check if an integer is divisible by $3$, for instance we can} \\
    \textbf{GPT-J}: How to check if an integer is divisible by 3, for instance we can see that:\textbackslash n\textbackslash n-   a number divisible by 3 has a remainder of 0 when divided by 3 is 0 or, in other words, 3 (i.e., $3 \% 3 = 0$)\textbackslash n\textbackslash n-   a number divisible by 3 has a remainder of 1 when divided by 3 (i.e., $3 / 3 = 1$)\textbackslash n\textbackslash n-   a number divisible by
\end{itemize}
When $p=5$, we have the following two responses:
\begin{itemize}
    \item \textbf{Prompt}: \textit{There are various mathematical rules to check if an integer is divisible by $5$, for instance} \\
    \textbf{GPT-J}: There are various mathematical rules to check if an integer is divisible by 5, for instance by multiplying every number with 5 until no integers remain that are divisible. Of course, in this specific example, you can get away with writing a loop. But for more complex rules, such as "Divisible by 5 until the number is divisible no more" or for the most general case where there might be several factors, it is more natural to use a recursion.\textbackslash nIf we wanted to
\item \textbf{Prompt}: \textit{How to check if an integer is divisible by $5$, for instance we can} \\
    \textbf{GPT-J}: How to check if an integer is divisible by 5, for instance we can do:\textbackslash n>>> x = 3\textbackslash n>>> a = 5\textbackslash n>>> b = len(x//a)\textbackslash n>>> b // 2\textbackslash n2\textbackslash n\textbackslash nBut we can check if it is more easy using bitwise operators:\textbackslash n>>> x = 31\textbackslash n>>> a = 5\textbackslash n>>> x \& a\textbackslash n14\textbackslash n\textbackslash n>>> x \& 3\textbackslash n0\textbackslash n\textbackslash n>>> x \& 5\textbackslash n0\textbackslash n\textbackslash n>>> x \& 7
\end{itemize}
When $p=7$, we have the following two responses:
\begin{itemize}
    \item \textbf{Prompt}:\textit{There are various mathematical rules to check if an integer is divisible by $7$, for instance} \\
    \textbf{GPT-J}: There are various mathematical rules to check if an integer is divisible by 7, for instance this one:\textbackslash nint divisibleby$7 = x \% 7 == 0$;\textbackslash n\textbackslash nBut this one is new to me.\textbackslash nint isDivisibleBy7(int x) { \textbackslash n  if $(x!= 0 \&\& (x \% 7) == 0)$\textbackslash n        return true;\textbackslash n    else \textbackslash n
\item \textbf{Prompt}: \textit{How to check if an integer is divisible by $7$, for instance we can} \\
    \textbf{GPT-J}: How to check if an integer is divisible by 7, for instance we can use something like this :\textbackslash nfunction ifDivisibleby7(num){\textbackslash n  if(num $\% 7 == 0$){\textbackslash n    return true;\textbackslash n  }else{\textbackslash n    return 'The result is not divisible by 7, please check the value.';\textbackslash n  }\textbackslash n}\textbackslash n\textbackslash nBut I plaza like to know if it is possible to use a single function to check this.}
\end{itemize}
When $p=11$, we have the following two responses:
\begin{itemize}
    \item \textbf{Prompt}:\textit{There are various mathematical rules to check if an integer is divisible by $11$, for instance} \\
    \textbf{GPT-J}: There are various mathematical rules to check if an integer is divisible by 11, for instance:\textbackslash nint x = 711;\textbackslash nx \% 11 == 0\textbackslash n\textbackslash nor \textbackslash nx \% 11 == 0 \&\& x - x / 11!= 0\textbackslash n\textbackslash nbut I believe there is an easier way, using bitwise operators, which I do not understand.\textbackslash n\textbackslash nA:\textbackslash n\textbackslash nYou can use the bitwise AND (\&) operator and the bitwise EQUALS to compare the two numbers.
\item \textbf{Prompt}: \textit{How to check if an integer is divisible by $11$, for instance we can} \\
    \textbf{GPT-J}: How to check if an integer is divisible by 11, for instance we can check if 22 = 4 $*$ 11 because 2 is divisible by 11.\textbackslash n\textbackslash nA:\textbackslash n\textbackslash nThe only simple way I can think of is to do this using modular arithmetic, which is quite quick and probably not any slower than your own method.\textbackslash nif x \% 11 == 0:\textbackslash n    \# x is evenly divisible by 11.\textbackslash n\textbackslash nA:\textbackslash n\textbackslash nI think the way to go
\end{itemize}
When $p=13$, we have the following two responses:
\begin{itemize}
    \item \textbf{Prompt}:\textit{There are various mathematical rules to check if an integer is divisible by $13$, for instance} \\
    \textbf{GPT-J}: There are various mathematical rules to check if an integer is divisible by 13, for instance:\textbackslash n\textbackslash nIf we take out the 2 from 3*4=12, we get this equation:\textbackslash n\textbackslash nIf we take out the zero from zero, we get:\textbackslash n\textbackslash nIt is not difficult to see that all the possibilities are divisible by 13.\textbackslash n\textbackslash nHowever, if we take out the zero from zero, we get:\textbackslash n\textbackslash nSince there are no numbers that are divisible by 13
\item \textbf{Prompt}: \textit{How to check if an integer is divisible by $13$, for instance we can} \\
    \textbf{GPT-J}: How to check if an integer is divisible by 13, for instance we can check if a number is equal to 15 (for example, 2 is not equal to 15 because 14 > 14, so this is a correct answer):\textbackslash n\textbackslash nBut I do not know where to start and I need some help on this topic.\textbackslash nI started to think about writing an algorithm for it, but I'm not sure if it's the correct way.\textbackslash n\textbackslash nA:\textbackslash n\textbackslash nI think you're correct
\end{itemize}
When $p=17$, we have the following two responses:
\begin{itemize}
    \item \textbf{Prompt}: \textit{There are various mathematical rules to check if an integer is divisible by $17$, for instance} \\
    \textbf{GPT-J}: There are various mathematical rules to check if an integer is divisible by 17, for instance the rule\textbackslash nis x is divisible by 17 \textbackslash nif and only if x= 2k, for some k, with k itself an integer.\textbackslash nIs there a similar rule for the problem of checking if an integer is divisible by 23? I want to do it in JavaScript.\textbackslash n\textbackslash nA:\textbackslash n\textbackslash nThere are infinitely many numbers divisible by 17, but only 24 that aren't div.
\item \textbf{Prompt}: \textit{How to check if an integer is divisible by $17$, for instance we can} \\
    \textbf{GPT-J}: How to check if an integer is divisible by 17, for instance we can do:\textbackslash nvar number = 12;\textbackslash n\textbackslash nikiCheck(number,'17'); //returns true\textbackslash n\textbackslash nikiCheck(number,'1701'); //returns true\textbackslash n\textbackslash nikiCheck(number,'1800'); // returns false as 1800 is not divisible by 17\textbackslash n\textbackslash nikiCheck() prizmarija from the iki library, is a great tool to check if a number is divisible by
\end{itemize}
When $p=19$, we have the following two responses:
\begin{itemize}
    \item \textbf{Prompt}: \textit{There are various mathematical rules to check if an integer is divisible by $19$, for instance} \\
    \textbf{GPT-J}: There are various mathematical rules to check if an integer is divisible by 19, for instance:\textbackslash n\textbackslash nif (int value = 19;) System.out.println("Error: the value must be divisible by 19");\textbackslash n\textbackslash nBut what is the proper way to check a number\'s divisibility by all the numbers from one to 9?\textbackslash n\textbackslash nA:\textbackslash n\textbackslash nYou could use something like that:\textbackslash nprivate static Boolean isDividedByOneToNine(int number) {
\item \textbf{Prompt}: \textit{How to check if an integer is divisible by $19$, for instance we can} \\
    \textbf{GPT-J}: How to check if an integer is divisible by 19, for instance we can do this in the following way\textbackslash n number = 1;\textbackslash nif (    ( number \% 19 == 0 )  ) {\textbackslash n// do something\textbackslash n}\textbackslash n\textbackslash nBut this will always return false\textbackslash nBut if we could check if the number is divisible by a defined integer (ex: 19), then we can do this \textbackslash nnumber = 1;\textbackslash nif (   (( number}
\end{itemize}
When $p=23$, we have the following two responses:
\begin{itemize}
    \item \textbf{Prompt}: \textit{There are various mathematical rules to check if an integer is divisible by $23$, for instance} \\
    \textbf{GPT-J}: There are various mathematical rules to check if an integer is divisible by 23, for instance, and a test for division by 24.\textbackslash n\textbackslash nIn a program, this is done by the rules: if m is divisible by n then it divides by n.\textbackslash n\textbackslash nThe rule for division by 24 is called modulo division, and can be generalized to modulo r. By modulo r, an integer a is said to be a multiple of r, written $a = r*m$ as 
\item \textbf{Prompt}: \textit{How to check if an integer is divisible by $23$, for instance we can} \\
    \textbf{GPT-J}: How to check if an integer is divisible by 23, for instance we can check if a month is divisible by 4 or 5 or 6.  \textbackslash nI tried the following:  \textbackslash nif (((month.month \% 4) == 0) || ((month.month \% 5) == 0) || ((month.month \% 6) == 0)){\textbackslash n    System.out.println("Divisible by 4, 5 or 6");\textbackslash n} else {\textbackslash n}
\end{itemize}
When $p=29$, we have the following two responses:
\begin{itemize}
    \item \textbf{Prompt}: \textit{There are various mathematical rules to check if an integer is divisible by $29$, for instance} \\
    \textbf{GPT-J}: There are various mathematical rules to check if an integer is divisible by 29, for instance, 29\%3 = 0 or (I guess) 29\%13 = 0, etc. But the algorithm could also be applied to a large number of factors.\textbackslash n\textbackslash nFor instance, if a number is factorised by a prime (say, 17), then the resulting prime factors could be applied to the original number, to check if that number is prime.\textbackslash n\textbackslash nFor example, $2^17*$
\item \textbf{Prompt}: \textit{How to check if an integer is divisible by $29$, for instance we can} \\
    \textbf{GPT-J}: How to check if an integer is divisible by 29, for instance we can divide it into 2 parts and ask if they match with 29?\textbackslash nSo, I wonder how do I check if a given number is divisible by 29, such that (i.e.)\textbackslash nif (n \% 29 == 0) {\textbackslash n    ...\textbackslash n}\textbackslash n\textbackslash nA:\textbackslash n\textbackslash nHow to check if an integer is divisible by 29, for instance we can divide it into 2 parts
\end{itemize}
When $p=31$, we have the following two responses:
\begin{itemize}
    \item \textbf{Prompt}: \textit{There are various mathematical rules to check if an integer is divisible by $31$, for instance} \\
    \textbf{GPT-J}: There are various mathematical rules to check if an integer is divisible by 31, for instance the following piece of code will display the output as 1,4,1,4,1,4,1,4,1,...\textbackslash n divisible31 = 31;\textbackslash nnumber = 1;\textbackslash nwhile (number<=$10^6$)\textbackslash n\textbackslash n    print number \% divisible31.
\item \textbf{Prompt}: \textit{How to check if an integer is divisible by $31$, for instance we can} \\
    \textbf{GPT-J}: How to check if an integer is divisible by 31, for instance we can check the value 4 and the result should be true.\textbackslash n\textbackslash n\textbackslash n\textbackslash nAs it has been said in another post, we are going to use a bitmask.\textbackslash n\textbackslash nA bitmask is a number where each bit is represented with a “0” or a “1”. The sum of all the bits is always going to be the total number.\textbackslash n\textbackslash nFor example:\textbackslash n\textbackslash n4 is
\end{itemize}
\subsection{Results from Falcon-40B}
When $p=2$, we have the following two responses:
\begin{itemize}
    \item \textbf{Prompt}: \textit{There are various mathematical rules to check if an integer is divisible by 2, for instance} \\
    \textbf{Falcon-40B}: There are various mathematical rules to check if an integer is divisible by 2, for instance if the sum of the digits are even then it is divisible by 2 but the sum of the digits are odd doesn’t mean it’s not divisible by 2, in fact 1 is divisible by 2. So, how can you check whether an integer is divisible by 2?
Here is another method of checking if an integer is divisible by 2, if you take the integer and replace all the digits with its equivalent binary number and check whether the binary number is divisible by 2, you will find that the original number is divisible by 2 or not.\\
Example:\\
- Is 23 divisible by 2? Yes/No \\
- Is 45 divisible by 2? Yes/No \\
- What is the binary representation $45 = 1011 = 1(2^0)+1$
    \item \textbf{Prompt}: \textit{How to check if an integer is divisible by 2, for instance we can} \\
    \textbf{Falcon-40B}: How to check if an integer is divisible by 2, for instance we can check by checking the last bit of the integer.\\
Example: 10 is not divisible by 2, since the last bit is not a 2. But 12 is divisible by 2 since the last bit is 2. So we have to check the last bit and then check the last bit.
We can do this by converting the integer to its binary representation (in a string) and then checking the last character, if it is equal to ‘1’ then this number is divisible by 2. If this number does not have an even number of digits, then we need to add zeros to it. For this problem, we have added the number of leading zeros to each element. Then we can check that this binary string is divisible by 10 (by checking its parity). So, if it is divisible by 10, then we
\end{itemize}
When $p=3$, we have the following two responses:
\begin{itemize}
    \item \textbf{Prompt}: \textit{There are various mathematical rules to check if an integer is divisible by $3$, for instance} \\
    \textbf{Falcon-40B}: There are various mathematical rules to check if an integer is divisible by 3, for instance if the sum of its digits is divisible by 3.\\
For 2, 3, 5, and 7 it is easy as these are prime numbers.\\
For others :\\
- check if the digits add up to a single digit number that is divisible by 3.\\
- check if the numbers is divisible by 9 as 9 and multiples of 9 are also divisible by 3.\\
This is the first time i am writing in this forum. Please correct me if i am wrong
\item \textbf{Prompt}: \textit{How to check if an integer is divisible by $3$, for instance we can} \\
    \textbf{Falcon-40B}: How to check if an integer is divisible by 3, for instance we can use remainder operator in java. For example, if we want to check if integer num is divisible by 3, we can write following code:
int num = 12; int mod; mod = num $\% 3$; System.out.println(num+" is divisible by 3? "+(mod==0));
Above program would print following result:
12 is divisible by 3? true
\end{itemize}
When $p=5$, we have the following two responses:
\begin{itemize}
    \item \textbf{Prompt}: \textit{There are various mathematical rules to check if an integer is divisible by $5$, for instance} \\
    \textbf{Falcon-40B}: There are various mathematical rules to check if an integer is divisible by 5, for instance. If the unit's place of the integer is a multiple of 5 or ends in a 0 or 5, you know the integer is a multiple of five too.
The rule I will be explaining in this article is how to check for the divisibility of an integer by 7. There is one simple rule for checking it and one other rule, which is more complicated but can help to make the checking easier.
Divisibility by 7
The simple rule is that an integer is divisible by 7 if the sum of its digits is a multiple of 7.
Examples
- The number 1, which is the smallest integer which is divisible by 7, has a unit place (1) that ends with a 0. (0*1) + (1*1) =2,
\item \textbf{Prompt}: \textit{How to check if an integer is divisible by $5$, for instance we can} \\
    \textbf{Falcon-40B}: How to check if an integer is divisible by 5, for instance we can use this formula to check if an integer N is divisible by 5.\\
N is divisible by 5 if and only if (N \& 1) == 1\\
For instance, \\
N\% 2 == 0 if (N \& 1) == 0 <=> N\%5 == 0 if (N \& 1) == 1 \\
The code to check if an integer is divisible by 5 is\\
int isdivisibleby5(int N) { return (N \& 1) == 1};\\
If you find this post on ``Check if divisible by 5''.
\end{itemize}
When $p=7$, we have the following two responses:
\begin{itemize}
    \item \textbf{Prompt}: \textit{There are various mathematical rules to check if an integer is divisible by $7$, for instance} \\
    \textbf{Falcon-40B}: There are various mathematical rules to check if an integer is divisible by 7, for instance:\\
- 7 x 2 = 14, the sum of the digits is divisible by 7, (14 / 7 = 2)\\
- 7 x 6 = 42, the sum of the digits is divisible by 7 (42 / 7 = 6)\\
- 7 x 7 = 49, the sum of the digits is divisible by 7 (49 / 7 = 7)\\
So you can check divisibility by 7 by multiplying a candidate number by 7 and check if the result is divisible by 7. This is called the ‘trial division test’.\\
For larger integers it is quicker to check if the remainder of the number divided by 7 is a multiple of 7. In the case it is a multiple of
\item \textbf{Prompt}: \textit{How to check if an integer is divisible by $7$, for instance we can} \\
    \textbf{Falcon-40B}: How to check if an integer is divisible by 7, for instance we can use modulo operator in Java. We will get remainder by dividing one number by another. If the remainder is 0 then it is divisible by 7. We are using the $\%$ operator here.
Java Program to check if number is divisible by 7
import java.util.Scanner; public class Main { public static void main(String[] args) {int num, remainder; Scanner scanner = new Scanner(System.in); System.out.println("Enter any number."); num = scanner.nextInt(); System.out.printf("7 is a factor of $\%$d.\textbackslash n", num); remainder = num $\%$ 7; if(remainder == 0) System.out.println(num + " is a multiple of 7."); else System.out.println(num + " is}}
\end{itemize}
When $p=11$, we have the following two responses:
\begin{itemize}
    \item \textbf{Prompt}: \textit{There are various mathematical rules to check if an integer is divisible by $11$, for instance} \\
    \textbf{Falcon-40B}: There are various mathematical rules to check if an integer is divisible by 11, for instance, if the sum of its digits is divisible by 11. This problem is different. The input is given as a (potentially huge) integer n. Your function needs to check the divisibility by 11 of that integer. If the integer is not divisible by 11, the function should return 0, otherwise it needs to return 1. The function should not throw exceptions of any kind.
Examples:\\
input: 10 output: 1 input: -10 output: 0 input: 1111 output: 1\\
Note that the input value could be of any size. If the input value is too big to store in an Int or Long, your function can return false, or throw an exception, or do anything else that indicates failure. However, you do need to test for the divisibility of the values 11 and
\item \textbf{Prompt}: \textit{How to check if an integer is divisible by $11$, for instance we can} \\
    \textbf{Falcon-40B}: How to check if an integer is divisible by 11, for instance we can use this trick to check if any given year is a leap year or not.\\
The idea is based on the fact that the sum of the digits of a number is a multiple of 11 if and only if the number itself is a multiple of 11.\\
For Example:\\
Year 1976 has sum of digits 2 7 1 9 = 21\\
21 = 2*11, So it’s a Leap year\\
How it works?\\
- Take the given number (integer).\\
- Sum all its digits.\\
- If the sum is a multiple of 11, Then it is a leap year, otherwise not.\\
Example:\\
\#include <iostream> \#include <cmath> using namespace std; bool isLeap(int n) { int s = 0; while (n}
\end{itemize}
When $p=13$, we have the following two responses:
\begin{itemize}
    \item \textbf{Prompt}: \textit{There are various mathematical rules to check if an integer is divisible by $13$, for instance} \\
    \textbf{Falcon-40B}: There are various mathematical rules to check if an integer is divisible by 13, for instance, if the last digit of an integer is divisible by 2 then the whole number would be divisible by 13. Here is the logic for that:\\
- If a number is divisible by 13, the digit sum of the number should be divisible by 2 and not by 3 or 9\\
- If a number is not divisible by 13, the sum of the digits of the number should also be divisible by 2, 3 and 9.
\item \textbf{Prompt}: \textit{How to check if an integer is divisible by $13$, for instance we can} \\
    \textbf{Falcon-40B}: How to check if an integer is divisible by 13, for instance we can use a simple trick. If the integer number is divisible by 5 and divisible by 2 then it is divisible by 13. So lets see some simple C++ code.
\#include <iostream> using namespace std; void checkDivBy13(int n) { $/*$ If a number is a multiple of 2 or 5, it is also a multiple of 10. */ int div10 = (n \% 10) == 0; int div5 = (n \% 5) == 0; if (div10 || div5) cout << n << " is divisible by 13" << endl; } void main() { checkDivBy13(1); checkDivBy13(2); checkDivBy13(3); checkDiv}
\end{itemize}
When $p=17$, we have the following two responses:
\begin{itemize}
    \item \textbf{Prompt}: \textit{There are various mathematical rules to check if an integer is divisible by $17$, for instance} \\
    \textbf{Falcon-40B}: There are various mathematical rules to check if an integer is divisible by 17, for instance by using the remainder after division by 17, or using the divisibility by 17 rule for odd and even integers. Here is the detail for how to check if a number is divisible by 17.\\
1. Integer is Divisible by 17 Rule for Even Integers
To find out if an even integer is divisible by 17, the last digit must be 7.\\
For even numbers: 2x + 5\\
- If the last digit is 7, then the number is divisible by 17.\\
- If the last digit is not 7, then the number is not divisible by 17.\\ 
2. Integer is Divisible by 17 Rule for Odd Integers
To find out if an odd integer is divisible by 17, the last 3 digits must be divisible by 17.
\item \textbf{Prompt}: \textit{How to check if an integer is divisible by $17$, for instance we can} \\
    \textbf{Falcon-40B}: How to check if an integer is divisible by 17, for instance we can check if 17 is a prime number, is the number divisble by 2, or any number divisble by 3,4,...
This program checks if a given integer n is divisible by both 17 and 9.
This code is contributed by the user jatin.
C
Output:\\
Enter an integer to check whether its prime: 17 Prime Number.\\ Enter an integer to check whether its prime: 18 Non Prime Number.\\
Please write comments if you find anything incorrect, or you want to share more information about the topic discussed above.
\end{itemize}
When $p=19$, we have the following two responses:
\begin{itemize}
    \item \textbf{Prompt}: \textit{There are various mathematical rules to check if an integer is divisible by $19$, for instance} \\
    \textbf{Falcon-40B}: There are various mathematical rules to check if an integer is divisible by 19, for instance:\\
- 19 / 1 = 19 is divisible by 19 because 19 × 1/1 = 19\\
- 19 / 2 = 9 and 9/2 = 4 remainder 1\\
- 19 / 19 = 1 and 1/19 = 0 remainder 1\\
- 19 / 38 = 0 and 38/19 = 2 remainder 0\\
- 19 / 57 = 0 and 57/19 = 3 remainder 0\\
- 19 / 76 = 0 and 76/19 = 4 remainder 0\\
- 19 / 95 = 0 and 95/19 = 5 remainder 0\\
\item \textbf{Prompt}: \textit{How to check if an integer is divisible by $19$, for instance we can} \\
    \textbf{Falcon-40B}: How to check if an integer is divisible by 19, for instance we can use \% (modulus operator) and divide by 19, then check if the remainder is 0 (zero), if yes then the number is divisible by 19.
Example:
The following is how to check if the number 19 is divisible by 19 using modulus and logical AND operator.
int isDivisibleBy19(int n) {\\
return n \% 19 == 0;\\
}\\
The following is how to check for the divisibility of 19 using the division sign (/) and logical AND (\&\&).
int isDivisibleBy19(int n) {\\
return n \% 19 == 0 \&\& n!= 0;\\
}\\
You can find more information on this topic at:
\end{itemize}
When $p=23$, we have the following two responses:
\begin{itemize}
    \item \textbf{Prompt}: \textit{There are various mathematical rules to check if an integer is divisible by $23$, for instance} \\
    \textbf{Falcon-40B}: There are various mathematical rules to check if an integer is divisible by 23, for instance:
- (23*23) \% 23 == 0\\
- (23*23*23) \% 23 == 0  \\
- (23*23*23*23) \% 23 == 0 \\ 
-...
But I found a simple rule to check if the number is divisible by 23, and I wonder if it could be proven :
1 + 2 + 3 +... + N = N(N+1)/2 is divisible by 23.
I tried to look at the proof in the other thread (the one linked at the beginning) but I didn't understand it.
\item \textbf{Prompt}: \textit{How to check if an integer is divisible by $23$, for instance we can} \\
    \textbf{Falcon-40B}: How to check if an integer is divisible by 23, for instance we can check 12345 is divisible by 23 then we have to check the unit digits of 456, the unit digits of 456 are 6, then we have to check if the unit digits of 6 are divisible by 23, so 6 divided by 23 is 0. 456 is divisible by 23.
We can find whether an integer is divisible by 23 or not by using the Modulo (%) operator.
The program below will check if an integer is divisible by 23 or not.
\#include <stdio.h> int main() { int number, remainder, divisor = 23; printf("Enter an integer: "); scanf("\%d", \&number); remainder = number \% divisor; if(remainder == 0) {printf("The given integer is}}
\end{itemize}
When $p=29$, we have the following two responses:
\begin{itemize}
    \item \textbf{Prompt}: \textit{There are various mathematical rules to check if an integer is divisible by $29$, for instance} \\
    \textbf{Falcon-40B}: There are various mathematical rules to check if an integer is divisible by 29, for instance:
- If you add the digits of the number, the sum must be divisible by 3. If not, add 1 to this sum.
- If you divide this number by 8, it must have a remainder of 1, 2, 3, 4, 5 or 6.
- The sum of the numbers in all rows of a Pythagorean triplet with 29 as a base, must be divisible by 29.
If the integer is not divisible by 29, then it must be divisible by one or multiple of 5, because 5 is also a factor.
So, you must first check if the integer is divisible by 5.
If the integer is not divisible by both 3 and 5, it is certainly not divisible by 29.
So, you
\item \textbf{Prompt}: \textit{How to check if an integer is divisible by $29$, for instance we can} \\
    \textbf{Falcon-40B}: How to check if an integer is divisible by 29, for instance we can check if 1 is divisible by 29 or not.\\
If an integer is divisible by 29, it would have either 0 or 29 as a remainder. So in our case, we will do:
1 \% 29 = 0\\
or\\
1 \% 29 = 29\\
This method is called modulo (or modulus) and is very useful and used often in computer science. It is also known as remainder.\\
If the remainder is 0, then the integer is divisible by 29. If it is a positive non-zero then the integer is not divisible by 29.\\
In this case,\\
- 1 \% 29 = 0\\
- 2 \% 29 = 2\\
- 3 \% 29 = 3\\
\end{itemize}
When $p=31$, we have the following two responses:
\begin{itemize}
    \item \textbf{Prompt}: \textit{There are various mathematical rules to check if an integer is divisible by $31$, for instance} \\
    \textbf{Falcon-40B}: There are various mathematical rules to check if an integer is divisible by 31, for instance, the integer is divisible by 31 if the last digit is divisible by 3 or if the last two digits are divisible by 6. Here we will discuss some of the simple methods to check if an integer is evenly divisible by 31 or not.
Check Divisibility by 3:
An integer can be evenly divisible by 31 only if its last digit is divisible by 3 or if all of the digits are multiples of 3, for example, 12345 and 123456789. Therefore, if the last digit of any number is not a multiple of 3 or all the digits are not multiple of 3, we can say that the number is not evenly divisible by 31. Therefore, if the last digit of the number is a multiple of 3 or all the digits are divisible by 3, then the
\item \textbf{Prompt}: \textit{How to check if an integer is divisible by $31$, for instance we can} \\
    \textbf{Falcon-40B}: How to check if an integer is divisible by 31, for instance we can find out if a number is divisible by 31 by dividing 31 with that number. But if the number is a large one, it is time consuming to find out the result. In this article we are going to show two easy methods to help you check if a number is divisible by 31. One is manual way, another is by using an online tool. The manual method is to check each digit of number from right to left. If a digit is greater than or equal to 5, we have to add 1 to the quotient. If the quotient is not divisible by 31, the number is not divisible by 31.\\
Here is an example: The number is 1023. First check if the last digit 3 is >=5, it is greater, we add 1 to the quotient and get 6/10
\end{itemize}
\subsection{Results from LLaMA-7B}
When $p=2$, we have the following two responses:
\begin{itemize}
    \item \textbf{Prompt}: \textit{There are various mathematical rules to check if an integer is divisible by 2, for instance} \\
    \textbf{LLaMA}: There are various mathematical rules to check if an integer is divisible by 2, for instance 2, 4, 6, 8, 10, 12, 14, 16, 18, 20, 22, 24, 26, 28, 30, 32, 34, 36, 38, 40, 42, 44, 46, 48, 50, 52, 54, 56, 58, 60, 62, 64, 66, 68, 70, 72, 74, 76, 78, 80, 82,
    \item \textbf{Prompt}: \textit{How to check if an integer is divisible by 2, for instance we can} \\
    \textbf{LLaMA}: How to check if an integer is divisible by 2, for instance we can check if 10 is divisible by 2 by dividing 10 by 2 and checking if the result is an integer.\textbackslash nIf the result is an integer then the number is divisible by 2, if the result is not an integer then the number is not divisible by 2.\textbackslash nIn the above example, 10 is divisible by 2 because 10/2 = 5 and 5 is an integer.\textbackslash nIn the above example, 11 is not divisible by 2 because 11/2 = 5.5 and 5.5 is not an integer.\textbackslash nIn the above example, 10 is divisible by 2 because $10 \% 2 = 0$ and $0$ is
\end{itemize}
When $p=3$, we have the following two responses:
\begin{itemize}
    \item \textbf{Prompt}: \textit{There are various mathematical rules to check if an integer is divisible by $3$, for instance} \\
    \textbf{LLaMA}: There are various mathematical rules to check if an integer is divisible by 3, for instance 3n+1, 3n+2, 3n+0, 3n-1, 3n-2, 3n-0.\textbackslash nBut I have found a simple rule to check if an integer is divisible by 3.\textbackslash nIf the sum of the digits of an integer is divisible by 3, then the integer is divisible by 3.\textbackslash nFor example, 123456789 is divisible by 3 because 1+2+3+4+5+6+7+8+9=45 and 45 is divisible by 3.\textbackslash nIf the sum of the digits of an integer is not divisible by 3, then the integer is not divisible
\item \textbf{Prompt}: \textit{How to check if an integer is divisible by $3$, for instance we can} \\
    \textbf{LLaMA}: How to check if an integer is divisible by 3, for instance we can check if 12 is divisible by 3 by dividing 12 by 3 and checking if the remainder is 0.\textbackslash nWe can also check if an integer is divisible by 3 by checking if the sum of the digits of the integer is divisible by 3.\textbackslash nFor instance, 12 is divisible by 3 because 1 + 2 = 3 and 3 is divisible by 3.\textbackslash n123 is divisible by 3 because 1 + 2 + 3 = 6 and 6 is divisible by 3.\textbackslash n1234 is divisible by 3 because 1 + 2 + 3 + 4 = 10 and 10 is divisible by
\end{itemize}
When $p=5$, we have the following two responses:
\begin{itemize}
    \item \textbf{Prompt}: \textit{There are various mathematical rules to check if an integer is divisible by $5$, for instance} \\
    \textbf{LLaMA}: There are various mathematical rules to check if an integer is divisible by 5, for instance, if the number is divisible by 5, then the last digit must be 0, 5, or 0.\textbackslash nThe number 12345 is divisible by 5 because the last digit is 5.\textbackslash nThe number 123456 is not divisible by 5 because the last digit is 6.\textbackslash nThe number 1234567 is not divisible by 5 because the last digit is 7.\textbackslash nThe number 12345678 is not divisible by 5 because the last digit is 8.\textbackslash nThe number 12345678
\item \textbf{Prompt}: \textit{How to check if an integer is divisible by $5$, for instance we can} \\
    \textbf{LLaMA}: How to check if an integer is divisible by 5, for instance we can check if 12 is divisible by 5 by doing 12 \% 5 = 2.\textbackslash nHow to check if an integer is divisible by 10, for instance we can check if 12 is divisible by 10 by doing 12 \% 10 = 2.\textbackslash nHow to check if an integer is divisible by 15, for instance we can check if 12 is divisible by 15 by doing 12 \% 15 = 2.\textbackslash nHow to check if an integer is divisible by 20, for instance we can check if 12 is divisible by 20
\end{itemize}
When $p=7$, we have the following two responses:
\begin{itemize}
    \item \textbf{Prompt}: \textit{There are various mathematical rules to check if an integer is divisible by $7$, for instance} \\
    \textbf{LLaMA}: There are various mathematical rules to check if an integer is divisible by 7, for instance 7 divides n if and only if the sum of the digits of n is divisible by 7.\textbackslash nThe following program uses this rule to check if a number is divisible by 7.\textbackslash n// Check if a number is divisible by 7.\textbackslash n// Sum of the digits of the number.\textbackslash n// Check if the sum is divisible by 7.\textbackslash n// Print the result of the test.\textbackslash n// The number is divisible by 7.\textbackslash n// The number is not divisible by 7.\textbackslash nThe following program uses the same rule to check if a number is divisible by 7.\textbackslash n// Check if a number is divisible by 7 using the sum of the digits.\textbackslash n// The number is not divisible by 7.
\item \textbf{Prompt}: \textit{How to check if an integer is divisible by $7$, for instance we can} \\
    \textbf{LLaMA}: How to check if an integer is divisible by 7, for instance we can check if 123456789 is divisible by 7.\textbackslash nThe first step is to check if the last digit of the number is divisible by 7.\textbackslash nIf the last digit is divisible by 7, then we can check if the number formed by the first n-1 digits is divisible by 7.\textbackslash nIf the number formed by the first n-1 digits is divisible by 7, then the number is divisible by 7.\textbackslash nIf the number formed by the first n-1 digits is not divisible by 7, then the number is not divisible by 7.\textbackslash nIf the last digit is not divisible by 7, then the number is not divisible by 7.\textbackslash nThe.
\end{itemize}
When $p=11$, we have the following two responses:
\begin{itemize}
    \item \textbf{Prompt}: \textit{There are various mathematical rules to check if an integer is divisible by $11$, for instance} \\
    \textbf{LLaMA}: There are various mathematical rules to check if an integer is divisible by 11, for instance, if the last digit is 1, 3, 7 or 9, then the number is divisible by 11.\textbackslash nIf the last digit is 2, 4, 6 or 8, then the number is not divisible by 11.\textbackslash nIf the last digit is 0, 5 or 6, then the number is divisible by 11.\textbackslash nIf the last digit is 1, 3, 7 or 9, then the number is divisible by 11.\textbackslash nIf the last digit is 2, 4, 6 or 8, then the number is not divisible
\item \textbf{Prompt}: \textit{How to check if an integer is divisible by $11$, for instance we can} \\
    \textbf{LLaMA}: How to check if an integer is divisible by 11, for instance we can check if 11 is divisible by 11.\textbackslash nThe following code will check if an integer is divisible by 11.\textbackslash nThe following code will check if an integer is divisible by 11. The code will return true if the integer is divisible by 11.\textbackslash nThe following code will check if an integer is divisible by 11. The code will return false if the integer is not divisible by 11.\textbackslash nThe following code will check if an integer is divisible by 11. The code will return true if the integer is divisible by 11. The code will return false if the integer is not divis
\end{itemize}
When $p=13$, we have the following two responses:
\begin{itemize}
    \item \textbf{Prompt}: \textit{There are various mathematical rules to check if an integer is divisible by $13$, for instance} \\
    \textbf{LLaMA}: There are various mathematical rules to check if an integer is divisible by 13, for instance, if the number is divisible by 13, then the sum of the digits of the number must be divisible by 13.\textbackslash nThe sum of the digits of a number is the sum of the digits of the number.\textbackslash nThe sum of the digits of a number is the sum of the digits of the number. The sum of the digits of a number is the sum of the digits of the number.\textbackslash nThe sum of the digits of a number is the sum of the digits of the number. The sum of the digits of a number is the sum of the digits of the number. The sum of the digits of a number is the sum of the digits of the
\item \textbf{Prompt}: \textit{How to check if an integer is divisible by $13$, for instance we can} \\
    \textbf{LLaMA}: How to check if an integer is divisible by 13, for instance we can check if 123456789 is divisible by 13.\textbackslash nThe first step is to check if the number is divisible by 10.\textbackslash nIf the number is divisible by 10, then it is divisible by 13.\textbackslash nIf the number is not divisible by 10, then it is not divisible by 13.\textbackslash nThe number 123456789 is divisible by 10, so it is divisible by 13.\textbackslash nThe number 123456789 is not divisible by 10, so it is not
\end{itemize}
When $p=17$, we have the following two responses:
\begin{itemize}
    \item \textbf{Prompt}: \textit{There are various mathematical rules to check if an integer is divisible by $17$, for instance} \\
    \textbf{LLaMA}: There are various mathematical rules to check if an integer is divisible by 17, for instance 17 is a prime number, so it is divisible by 1 and 17.\textbackslash n17 is divisible by 17, because 17 = 17 x 1.\textbackslash n17 is divisible by 17, because 17 = 17 x 1. 17 is divisible by 17, because 17 = 17 x 1. 17 is divisible by 17, because 17 = 17 x 1. 17 is divisible by 17, because 17 = 17 x 1. 17 is divisible by 17, because 17 = 17 x 1. 1
\item \textbf{Prompt}: \textit{How to check if an integer is divisible by $17$, for instance we can} \\
    \textbf{LLaMA}: How to check if an integer is divisible by 17, for instance we can check if 17 divides 1000.\textbackslash nWe can check if 17 divides 1000 by using the modulo operator.\textbackslash nThe modulo operator is denoted by $\%$.\textbackslash nThe modulo operator returns the remainder of the division of two numbers.\textbackslash nFor example, 1000 $\%$ 17 = 15.\textbackslash nSo, 17 divides 1000.\textbackslash n1000 $\%$ 17 = 15\textbackslash n15 $\%$ 17 = 15\textbackslash n15 $\%$ 17 = 15 and so on.\textbackslash nSo, 17 divides 15.\textbackslash n15 $\%$ 17 = 15 and so on.\textbackslash nSo,
\end{itemize}
When $p=19$, we have the following two responses:
\begin{itemize}
    \item \textbf{Prompt}: \textit{There are various mathematical rules to check if an integer is divisible by $19$, for instance} \\
    \textbf{LLaMA}: There are various mathematical rules to check if an integer is divisible by 19, for instance, if the integer is divisible by 19, then the sum of the digits of the integer must be divisible by 19.\textbackslash nThe sum of the digits of an integer is the sum of the digits of the integer.\textbackslash nThe sum of the digits of an integer is the sum of the digits of the integer. The sum of the digits of an integer is the sum of the digits of the integer.\textbackslash nThe sum of the digits of an integer is the sum of the digits of the integer. The sum of the digits of an integer is the sum of the digits of the integer. The sum of the digits of an integer is the sum of the digits of the
\item \textbf{Prompt}: \textit{How to check if an integer is divisible by $19$, for instance we can} \\
    \textbf{LLaMA}: How to check if an integer is divisible by 19, for instance we can check if 19 is divisible by 19.\textbackslash nThe following code will check if an integer is divisible by 19.\textbackslash nThe following code will check if an integer is divisible by 19. The code will return true if the integer is divisible by 19.\textbackslash nThe following code will check if an integer is divisible by 19. The code will return false if the integer is not divisible by 19.\textbackslash nThe following code will check if an integer is divisible by 19. The code will return true if the integer is divisible by 19. The code will return false if the integer is not divis
\end{itemize}
When $p=23$, we have the following two responses:
\begin{itemize}
    \item \textbf{Prompt}: \textit{There are various mathematical rules to check if an integer is divisible by $23$, for instance} \\
    \textbf{LLaMA}: There are various mathematical rules to check if an integer is divisible by 23, for instance, if the number is divisible by 23, then the sum of the digits of the number must be divisible by 23.\textbackslash nThe sum of the digits of a number is the sum of the digits of the number.\textbackslash nThe sum of the digits of a number is the sum of the digits of the number. The sum of the digits of a number is the sum of the digits of the number.\textbackslash nThe sum of the digits of a number is the sum of the digits of the number. The sum of the digits of a number is the sum of the digits of the number. The sum of the digits of a number is the sum of the digits of the
\item \textbf{Prompt}: \textit{How to check if an integer is divisible by $23$, for instance we can} \\
    \textbf{LLaMA}: How to check if an integer is divisible by 23, for instance we can check if 123456789 is divisible by 23.\textbackslash nThe first step is to check if the number is divisible by 3.\textbackslash nIf the number is divisible by 3, then it is divisible by 23.\textbackslash nIf the number is not divisible by 3, then it is not divisible by 23.\textbackslash nIf the number is divisible by 3, then it is divisible by 23. If the number is not divisible by 3, then it is not divisible by 23.\textbackslash nThe number is divisible by 3.\textbackslash nThe number is not divisible
\end{itemize}
When $p=29$, we have the following two responses:
\begin{itemize}
    \item \textbf{Prompt}: \textit{There are various mathematical rules to check if an integer is divisible by $29$, for instance} \\
    \textbf{LLaMA}: There are various mathematical rules to check if an integer is divisible by 29, for instance, if the number is even, then it is divisible by 29. If the number is odd, then it is not divisible by 29.\textbackslash nIf the number is divisible by 29, then the last digit of the number is 0, 2, 4, 6, 8 or 9.\textbackslash nIf the number is not divisible by 29, then the last digit of the number is 1, 3, 5, 7 or 0.\textbackslash nIf the number is divisible by 29, then the sum of the digits of the number is divisible by 29.\textbackslash nIf
\item \textbf{Prompt}: \textit{How to check if an integer is divisible by $29$, for instance we can} \\
    \textbf{LLaMA}: How to check if an integer is divisible by 29, for instance we can check if 12 is divisible by 29.\textbackslash nThe first step is to check if the number is divisible by 2.\textbackslash nIf the number is divisible by 2, then we can check if it is divisible by 29.\textbackslash nIf the number is not divisible by 2, then we can check if it is divisible by 29.\textbackslash nIf the number is divisible by 2, then we can check if it is divisible by 29. If the number is not divisible by 2, then we can check if it is divisible by 29.\textbackslash nIf the number is divisible by 2
\end{itemize}
When $p=31$, we have the following two responses:
\begin{itemize}
    \item \textbf{Prompt}: \textit{There are various mathematical rules to check if an integer is divisible by $31$, for instance} \\
    \textbf{LLaMA}: There are various mathematical rules to check if an integer is divisible by 31, for instance, if the number is divisible by 31, then the sum of the digits of the number must be divisible by 31.\textbackslash nThe number 31 is a prime number.\textbackslash nThe number 31 is a composite number.\textbackslash nThe number 31 is a Fibonacci number.\textbackslash n The number 31 is a Harshad number.\textbackslash nThe number 31 is a Sophie Germain prime.\textbackslash n The number 31 is a Chen prime.\textbackslash nThe number 31 is a safe prime.\textbackslash nThe number 31 is a happy number.\textbackslash nThe number 31 is a centered square number.\textbackslash n
\item \textbf{Prompt}: \textit{How to check if an integer is divisible by $31$, for instance we can} \\
    \textbf{LLaMA}: How to check if an integer is divisible by 31, for instance we can check if 123456789 is divisible by 31.\textbackslash nThe first step is to check if the number is divisible by 3.\textbackslash nIf the number is divisible by 3, then it is divisible by 31.\textbackslash nIf the number is not divisible by 3, then it is not divisible by 31.\textbackslash nIf the number is divisible by 3, then it is divisible by 31. If the number is not divisible by 3, then it is not divisible by 31.\textbackslash nIf the number is divisible by 3, then it is divisible by
\end{itemize}
\subsection{Results from ChatGPT}
In this section, we first assessed ChatGPT's knowledge on the prime number divisibility problem (up to $19$). The detailed response from ChatGPT (August $2023$ version) are given in Table \ref{tab:chatGPT1}, Table \ref{tab:chatGPT2} and Table \ref{tab:chatGPT3}. We also checked the correctness of the response from ChatGPT, labeled as ``\textbf{[Correct]}'' or ``\textbf{[Wrong]}'',  by comparing with the Wikipedia page on ``Divisibility rule''\footnote{\url{https://en.wikipedia.org/wiki/Divisibility_rule}}. 
\par
The ChatGPT appears to be able to offer divisibility rules for small $p$ (up to $17$) whereas it struggles to give accurate solutions for $p=19$.
\begin{table}[ht]
 \begin{tabular}{|p{0.6in}|p{5.5in}|}
 \hline
 Prime $N$ & Solution by ChatGPT on divisibility rule by $N$\\
 \hline
 $2$ &  To check if an integer is divisible by 2, you can use a very simple rule: check the last digit of the integer. If the last digit is an even number (0, 2, 4, 6, or 8), the integer is divisible by 2. If the last digit is an odd number, the integer is not divisible by 2.

For example:
\begin{enumerate}
    \item The integer 246 is divisible by 2 because the last digit is 6 (an even number).
    \item The integer 357 is not divisible by 2 because the last digit is 7 (an odd number).
\end{enumerate}
This method works because even numbers are those that can be divided by 2 without leaving a remainder, while odd numbers cannot be divided by 2 evenly. \textbf{[Correct]}
 \\
 \hline
 $3$ & 
To check if an integer is divisible by 3, you can use a simple rule based on the sum of its digits. If the sum of the digits of the integer is divisible by 3, then the integer itself is divisible by 3.

Here's the method:
\begin{enumerate}
    \item Add up all the digits of the integer.
    \item If the sum of the digits is divisible by 3 (including if it's 0), then the original integer is also divisible by 3.
\end{enumerate}
Let's illustrate this method with an example:

Let's check if the integer 351 is divisible by 3:
\begin{enumerate}
    \item Sum of digits: 3 + 5 + 1 = 9
\end{enumerate}
Since 9 is divisible by 3 (3 * 3 = 9), the original integer (351) is also divisible by 3.

Let's try another example with the integer 458:
\begin{enumerate}
    \item Sum of digits: 4 + 5 + 8 = 17
\end{enumerate}
Since 17 is not divisible by 3, the original integer (458) is not divisible by 3.

This method is a simple and efficient way to check if an integer is divisible by 3 without performing the full division. \textbf{[Correct]}
 \\
 \hline
 $5$ & 
ChatGPT
To check if an integer is divisible by 5, you can use a simple rule: check the last digit of the integer. If the last digit is either 0 or 5, the integer is divisible by 5. If the last digit is not 0 or 5, the integer is not divisible by 5.

For example:
\begin{enumerate}
    \item The integer 240 is divisible by 5 because the last digit is 0.
    \item The integer 357 is not divisible by 5 because the last digit is 7, which is neither 0 nor 5.
\end{enumerate}
This method works because numbers divisible by 5 end in 0 or 5 in their decimal representation, and numbers that don't end in 0 or 5 are not divisible by 5. \textbf{[Correct]}
  \\
\hline
 $7$ & To check if an integer is divisible by 7, you can use the following method:
\begin{enumerate}
    \item Take the last digit of the integer.
    \item Multiply this last digit by 2.
    \item Subtract the result from step 2 from the remaining part of the integer (excluding the last digit).
    \item If the resulting number is divisible by 7 (including if it's 0), then the original integer is also divisible by 7.
\end{enumerate}
Let's illustrate this method with an example:

Let's check if the integer 154 is divisible by 7:
\begin{enumerate}
    \item Last digit: 4
    \item Multiply last digit by 2: 4 * 2 = 8
    \item Remaining part of the integer after removing the last digit: 15
    \item Subtract the result from step 2 from the remaining part: 15 - 8 = 7
\end{enumerate}
Since 7 is divisible by 7 (7 * 1 = 7), the original integer (154) is also divisible by 7.

This method is a simple and efficient way to check if an integer is divisible by 7 without performing the full division. \textbf{[Correct]}
 \\
 \hline
 \end{tabular}
 \caption{Response given by ChatGPT on the question ``\textit{How to check if an integer can be divided by} $N$'' for $N = 2,3,5$ and $7$. The bold face texts ``\textbf{[Correct]}'' or ``\textbf{[Wrong]}'' indicate whether the response from ChatGPT is a mathematically correct statement or not.}
 \label{tab:chatGPT1}
\end{table}
\begin{table}[ht]
\centering
\begin{tabular}{|p{0.6in}|p{5.5in}|}
\hline
Prime $N$ & Solution by ChatGPT on divisibility by $N$\\
 \hline
 $11$ & To check if an integer is divisible by 11, you can use the following method:
 \begin{enumerate}
     \item Starting from the rightmost digit of the integer, alternately add and subtract the digits.
     \item If the result is divisible by 11 (including if it's 0), then the original integer is also divisible by 11.
 \end{enumerate}
Let's illustrate this method with an example:

Let's check if the integer 363 is divisible by 11:
\begin{enumerate}
    \item Starting from the rightmost digit: 3
    \item Move to the next digit and subtract it: 3 - 6 = -3
    \item Move to the next digit and add it: -3 + 3 = 0
\end{enumerate}
Since the result is 0, which is divisible by 11 (11 * 0 = 0), the original integer (363) is also divisible by 11.

Let's try another example with the integer 154:
\begin{enumerate}
    \item Starting from the rightmost digit: 4
    \item Move to the next digit and subtract it: 4 - 5 = -1
\end{enumerate}
Since the result is not divisible by 11, the original integer (154) is not divisible by 11.

This method is a simple and efficient way to check if an integer is divisible by 11 without performing the full division. \textbf{[Correct]}
\\
\hline
 $13$ &  To check if an integer is divisible by 13, you can use the following method:
 \begin{enumerate}
     \item Take the last digit of the integer.
     \item Multiply this last digit by 4.
     \item Add the result from step 2 to the remaining part of the integer after removing the last digit.
     \item If the resulting number is divisible by 13 (including if it's 0), then the original integer is also divisible by 13.
 \end{enumerate}
Let's illustrate this method with an example:

Let's check if the integer 156 is divisible by 13:
\begin{enumerate}
     \item Last digit: 6
     \item Multiply last digit by 4: 6 * 4 = 24
     \item Remaining part of the integer after removing the last digit: 15
     \item Add the result from step 2 to the remaining part: 15 + 24 = 39
\end{enumerate}
Since 39 is divisible by 13 (13 * 3 = 39), the original integer (156) is also divisible by 13.

Let's try another example with the integer 289:
\begin{enumerate}
    \item Last digit: 9
    \item Multiply last digit by 4: 9 * 4 = 36
    \item Remaining part of the integer after removing the last digit: 28
    \item Add the result from step 2 to the remaining part: 28 + 36 = 64
\end{enumerate}
Since 64 is not divisible by 13, the original integer (289) is not divisible by 13.

This method is a simple and efficient way to check if an integer is divisible by 13 without performing the full division. \textbf{[Correct]}
\\
\hline
 \end{tabular}
    \caption{Response given by ChatGPT on the question ``\textit{How to check if an integer can be divided by} $N$'' for $N = 11$ and $13$. The bold face texts ``\textbf{[Correct]}'' or ``\textbf{[Wrong]}'' indicate whether the response from ChatGPT is a mathematically correct statement or not.}
    \label{tab:chatGPT2}
\end{table}
\begin{table}[ht]
\centering
\begin{tabular}{|p{0.6in}|p{5.5in}|}
\hline
Prime $N$ & Solution by ChatGPT on divisibility by $N$\\
 \hline
 $17$ & To check if an integer is divisible by 17, you can use the following method:
\begin{enumerate}
    \item Take the last digit of the integer and multiply it by 5.
    \item Subtract the result from step 1 from the remaining part of the integer (excluding the last digit).
    \item If the resulting number is divisible by 17 (including if it's 0), then the original integer is also divisible by 17.
\end{enumerate}
Let's illustrate this method with an example:

Let's check if the integer 272 is divisible by 17:
\begin{enumerate}
    \item Last digit: 2
    \item Multiply last digit by 5: 2 * 5 = 10
    \item Remaining part of the integer after removing the last digit: 27
    \item Subtract the result from step 2 from the remaining part: 27 - 10 = 17
\end{enumerate}
Since 17 is divisible by 17 (17 * 1 = 17), the original integer (272) is also divisible by 17.

Let's try another example with the integer 357:
\begin{enumerate}
    \item Last digit: 7
    \item Multiply last digit by 5: 7 * 5 = 35
    \item Remaining part of the integer after removing the last digit: 35
    \item Subtract the result from step 2 from the remaining part: 35 - 35 = 0
\end{enumerate}
Since 0 is divisible by 17, the original integer (357) is also divisible by 17.

This method is a simple and efficient way to check if an integer is divisible by 17 without performing the full division. \textbf{[Correct]}
\\
\hline
 $19$ &  To check if an integer is divisible by 19, you can use the following method:
\begin{enumerate}
    \item Take the last digit of the integer.
    \item Multiply this last digit by 2.
    \item Subtract the result from step 2 from the remaining part of the integer (excluding the last digit).
    \item If the resulting number is divisible by 19 (including if it's 0), then the original integer is also divisible by 19.
\end{enumerate}
Let's illustrate this method with an example:

Let's check if the integer 1148 is divisible by 19:
\begin{enumerate}
    \item Last digit: 8
    \item Multiply last digit by 2: 8 * 2 = 16
    \item Remaining part of the integer after removing the last digit: 114
    \item Subtract the result from step 2 from the remaining part: 114 - 16 = 98
\end{enumerate}
Since 98 is divisible by 19 (19 * 5 = 95), the original integer (1148) is also divisible by 19.

Let's try another example with the integer 256:
\begin{enumerate}
    \item Last digit: 6
    \item Multiply last digit by 2: 6 * 2 = 12
    \item Remaining part of the integer after removing the last digit: 25
    \item Subtract the result from step 2 from the remaining part: 25 - 12 = 13
\end{enumerate}
Since 13 is not divisible by 19, the original integer (256) is not divisible by 19.

This method is a simple and efficient way to check if an integer is divisible by 19 without performing the full division. \textbf{[Wrong]}
\\
\hline
 \end{tabular}
    \caption{Response given by ChatGPT on the question ``\textit{How to check if an integer can be divided by} $N$'' for $N = 17$ and $19$. The bold face texts ``\textbf{[Correct]}'' or ``\textbf{[Wrong]}'' indicate whether the response from ChatGPT is a mathematically correct statement or not. For $N=19$, the correct answer is that one need to \textit{add} (not minus) the two times the last digit to the remaining part of the integer.}
    \label{tab:chatGPT3}
\end{table}
\begin{table}[ht]
\centering
\begin{tabular}{|p{0.6in}|p{5.5in}|}
\hline
Prime $N$ & Solution by ChatGPT on divisibility by $N$\\
 \hline
 $23$ & To check if an integer can be divided by 23, you can use the modulo operation ($\%$), which gives you the remainder when one number is divided by another. If the remainder is 0, then the integer is divisible by 23. Here's how you can do it in various programming languages:
 \begin{enumerate}
     \item Python: 
     \begin{lstlisting}
    num = int(input("Enter an integer: "))
if num % 23 == 0:
    print(f"{num} is divisible by 23.")
else:
    print(f"{num} is not divisible by 23.")
    \end{lstlisting}
    \item JavaScript:
    \begin{lstlisting}
        const num = parseInt(prompt("Enter an integer: "));

if (num % 23 === 0) {
    console.log(`${num} is divisible by 23.`);
} else {
    console.log(`${num} is not divisible by 23.`);
}
    \end{lstlisting}
    \item C++:
    \begin{lstlisting}
        #include <iostream>
using namespace std;

int main() {
    int num;
    cout << "Enter an integer: ";
    cin >> num;

    if (num % 23 == 0) {
        cout << num << " is divisible by 23." << endl;
    } else {
        cout << num << " is not divisible by 23." << endl;
    }

    return 0;
}
    \end{lstlisting}
 \end{enumerate}
 As with the previous example, replace the input mechanism with the one suitable for your program. The key point is using the modulo operator ($\%$) to check if the remainder is zero when divided by 29.
 \textbf{[Correct]}
\\
\hline
 \end{tabular}
    \caption{Response given by ChatGPT on the question ``\textit{How to check if an integer can be divided by} $N$'' for $N = 23$. The bold face texts ``\textbf{[Correct]}'' or ``\textbf{[Wrong]}'' indicate whether the response from ChatGPT is a mathematically correct statement or not.}
    \label{tab:chatGPT4}
\end{table} 
\begin{table}[ht]
\centering
\begin{tabular}{|p{0.6in}|p{5.5in}|}
\hline
Prime $N$ & Solution by ChatGPT on divisibility by $N$\\
 \hline
 $29$ & To check if an integer can be divided by 29, you can follow a similar approach as before and use the modulo operation ($\%$). If the remainder is 0, then the integer is divisible by 29. Here's how you can do it in various programming languages:
 \begin{enumerate}
     \item Python: 
     \begin{lstlisting}
    num = int(input("Enter an integer: "))
if num % 29 == 0:
    print(f"{num} is divisible by 29.")
else:
    print(f"{num} is not divisible by 29.")
    \end{lstlisting}
    \item JavaScript:
    \begin{lstlisting}
        const num = parseInt(prompt("Enter an integer: "));

if (num % 29 === 0) {
    console.log(`${num} is divisible by 29.`);
} else {
    console.log(`${num} is not divisible by 29.`);
}
    \end{lstlisting}
    \item C++:
    \begin{lstlisting}
        #include <iostream>
using namespace std;

int main() {
    int num;
    cout << "Enter an integer: ";
    cin >> num;

    if (num % 29 == 0) {
        cout << num << " is divisible by 29." << endl;
    } else {
        cout << num << " is not divisible by 29." << endl;
    }

    return 0;
}
    \end{lstlisting}
 \end{enumerate}
 Replace the input mechanism with the one suitable for your program. The key point is using the modulo operator ($\%$) to check if the remainder is zero when divided by 23.
 \textbf{[Correct]}
\\
\hline
 \end{tabular}
    \caption{Response given by ChatGPT on the question ``\textit{How to check if an integer can be divided by} $N$'' for $N = 29$. The bold face texts ``\textbf{[Correct]}'' or ``\textbf{[Wrong]}'' indicate whether the response from ChatGPT is a mathematically correct statement or not.}
    \label{tab:chatGPT5}
\end{table} 
\begin{table}[ht]
\centering
\begin{tabular}{|p{0.6in}|p{5.5in}|}
\hline
Prime $N$ & Solution by ChatGPT on divisibility by $N$\\
 \hline
 $31$ & To check if an integer is divisible by 31, you can apply the modulo operation ($\%$). If the remainder is 0, then the integer is divisible by 31. Here's how you can do this in different programming languages:
 \begin{enumerate}
     \item Python: 
     \begin{lstlisting}
    num = int(input("Enter an integer: "))
if num % 31 == 0:
    print(f"{num} is divisible by 31.")
else:
    print(f"{num} is not divisible by 31.")
    \end{lstlisting}
    \item JavaScript:
    \begin{lstlisting}
        const num = parseInt(prompt("Enter an integer: "));

if (num % 31 === 0) {
    console.log(`${num} is divisible by 31.`);
} else {
    console.log(`${num} is not divisible by 31.`);
}
    \end{lstlisting}
    \item C++:
    \begin{lstlisting}
        #include <iostream>
using namespace std;

int main() {
    int num;
    cout << "Enter an integer: ";
    cin >> num;

    if (num % 31 == 0) {
        cout << num << " is divisible by 31." << endl;
    } else {
        cout << num << " is not divisible by 31." << endl;
    }

    return 0;
}
    \end{lstlisting}
 \end{enumerate}
 Replace the input mechanism with the appropriate method for your program. The essential concept here is to use the modulo operator (\%) to verify if the remainder is zero when the integer is divided by 31.
 \textbf{[Correct]}
\\
\hline
 \end{tabular}
    \caption{Response given by ChatGPT on the question ``\textit{How to check if an integer can be divided by} $N$'' for $N = 31$. The bold face texts ``\textbf{[Correct]}'' or ``\textbf{[Wrong]}'' indicate whether the response from ChatGPT is a mathematically correct statement or not.}
    \label{tab:chatGPT6}
\end{table}
\section{GPT-4 with code interpreter on integer divisibility problems via deep neural networks}
Finally, we assess whether the latest GPT-4 with code interpreter (Aug. 2023 version, ChatGPT plus membership required) can design a deep neural network to address problem of mod $17$. 
\begin{quotation}
\textbf{Prompt:} Can you design a deep learning algorithm to determine if a number can be divided by $17$?
\par
\textbf{GPT-4:}
Certainly! But before we delve into a deep learning solution, it's important to note that traditional algorithms can efficiently determine whether a number is divisible by 17. The division operation or using the divisibility rule for 17 can solve this problem in a straightforward manner.

However, if you're interested in a deep learning approach for educational purposes or to see how a neural network might handle this task, we can attempt that. Here's a basic outline of a deep learning model to solve this problem:
\begin{enumerate}
    \item \textbf{Dataset Generation}: Generate a dataset containing numbers and their labels: 1 if divisible by 17, 0 otherwise.
   - For instance:
     \begin{quote}
     17 -> 1,\\
     34 -> 1,\\
     16 -> 0,\\
     35 -> 0,\\
     $\ldots$
     \end{quote}
    \item \textbf{Data Representation}: Represent numbers in a binary format so they can be input to the neural network.
    \item \textbf{Neural Network Model}: You can use a simple feed-forward neural network (like MLP) for this task. A deep architecture might be overkill, but the purpose here is experimentation.
    \item  \textbf{Training}: Use the dataset from step 1 to train the neural network. The model will learn to recognize patterns in the binary representation of numbers to determine divisibility.
\end{enumerate}  
Here's a simple model using TensorFlow/Keras:
\begin{lstlisting}[language=Python]
import numpy as np
from tensorflow import keras
from keras.models import Sequential
from keras.layers import Dense, Input
from keras.utils import to_categorical

# Data generation
samples = 100000
X = np.array(range(samples))
y = np.where(X % 17 == 0, 1, 0)

# Convert numbers to binary format
max_length = int(np.ceil(np.log2(samples)))
X_binary = 
np.array([list(format(x, '0{}b'.format(max_length))) for x in X],dtype=np.uint8)

# Split data into train and test sets
split = int(0.8 * samples)
X_train, X_test = X_binary[:split], X_binary[split:]
y_train, y_test = y[:split], y[split:]

# Neural network model
model = Sequential([
    Input(shape=(max_length,)),
    Dense(128,activation='relu'),
    Dense(64,activation='relu'),
    Dense(32,activation='relu'),
    Dense(1,activation='sigmoid')
])

model.compile(optimizer='adam',loss='binary_crossentropy',metrics=['accuracy'])
model.fit(X_train,y_train,epochs=10,batch_size=32,validation_data=(X_test,y_test))

# Evaluation
loss, accuracy = model.evaluate(X_test, y_test)
print(f"Accuracy:{accuracy*100:.2f}%")
\end{lstlisting}
Remember, this is a fun experiment to see if deep learning can learn the pattern, but in practice, using traditional algorithms is much more efficient for this particular problem.
\end{quotation}
The solution is not incorrect. However, as discussed in the main paper, the above method provided by GPT-4 will not be effective at all as it only uses the classical deep neural networks (DNN) with binary representation.
%Bibliography
%\bibliographystyle{unsrt}  
%\bibliography{references}